\documentclass[10pt,twocolumn,letterpaper]{article}

\usepackage{iccv}
\usepackage{times}
\usepackage{epsfig}
\usepackage{graphicx}
\usepackage{amsmath}
\usepackage{amssymb}

\makeatletter
\@namedef{ver@everyshi.sty}{}
\makeatother

\usepackage[utf8]{inputenc} %
\usepackage[T1]{fontenc}    %
\usepackage{url}            %
\usepackage{booktabs}       %
\usepackage{amsfonts}       %
\usepackage{nicefrac}       %
\usepackage{microtype}      %
\usepackage{pgf-pie}

\usepackage[square,numbers,sort&compress]{natbib}
\usepackage{enumitem}
\usepackage{algpseudocode}
\usepackage{xcolor}
\usepackage{array}
\usepackage{multirow}
\usepackage{algorithm}
\usepackage[font=small,labelfont=bf]{caption}
\usepackage{subcaption}
\usepackage[normalem]{ulem}
\usepackage{xparse}
\usepackage{pifont}
\usepackage{bm}
\usepackage{threeparttable}
\usepackage{etoolbox}
\usepackage{listings}
\usepackage{mwe} %
\usepackage{makecell}
\usepackage{color, colortbl}
\usepackage{tabularx}
\usepackage[accsupp]{axessibility}
\usepackage{soul}
\usepackage{tabu}
\usepackage{makecell}
\usepackage[scaled=0.85]{DejaVuSansMono}
\usepackage{tikz}
\usepackage{pgfplots}
\pgfplotsset{compat=1.16}
\usepgfplotslibrary{groupplots}
\usetikzlibrary{patterns}
\usetikzlibrary{arrows}
\usepackage{tkz-kiviat}
\usepackage{xfp}

\definecolor{citecolor}{HTML}{0071bc}
\usepackage[breaklinks=true,bookmarks=false,colorlinks,bookmarks=false, citecolor=citecolor]{hyperref}

\usepackage[capitalize]{cleveref}
\crefname{section}{\S}{\S\S}
\crefname{subsection}{\S}{\S\S}
\crefformat{table}{Table~#2#1#3}
\crefformat{figure}{Figure~#2#1#3}
\crefformat{equation}{Eq~#2#1#3}

\makeatletter
\DeclareRobustCommand\onedot{\futurelet\@let@token\@onedot}
\def\@onedot{\ifx\@let@token.\else.\null\fi\xspace}

\def\ie{\emph{i.e}\onedot} 
 
\def\etc{\emph{etc}\onedot} \def\vs{\emph{vs}\onedot}

\def\etal{\emph{et al}\onedot}

\makeatother

\newlength\savewidth

\newlength\thinwidth

\newif\ifarxiv
\arxivfalse

\definecolor{Gray}{gray}{0.92}
\definecolor{DarkGray}{gray}{0.5}
\newcolumntype{a}{>{\columncolor{Gray}}c}
\newcolumntype{H}{>{\setbox0=\hbox\bgroup}c<{\egroup}@{}}
\definecolor{LightCyan}{rgb}{0.88,1,1}
\definecolor{altRowColor}{gray}{0.92}
\definecolor{highlightRowColor}{gray}{0.9}

\DeclareRobustCommand{\colorrowtext}[0]{{\sethlcolor{highlightRowColor}\hl{gray}}}
\soulregister{\colorrowtext}{1}

\definecolor{GrayNumber}{gray}{0.5}
\definecolor{GrayXMark}{gray}{0.7}

\definecolor{ImageDark}{rgb}{0,0.3,0.8}
\definecolor{VideoDark}{rgb}{.5,.0,.5}
\definecolor{LowShotDark}{rgb}{.5,0.1,0}
\definecolor{ZeroShotDark}{rgb}{0.8,0.5,0.2}
\definecolor{LinearDark}{rgb}{0.5,0.5,0}
\definecolor{DetDark}{rgb}{0.0,0.5,0.0}
\colorlet{Image}{ImageDark!20!white}
\colorlet{Video}{VideoDark!20!white}
\colorlet{Det}{DetDark!20!white}
\colorlet{ImageLight}{ImageDark!70!white}
\colorlet{VideoLight}{VideoDark!70!white}
\colorlet{DetLight}{DetDark!70!white}
\definecolor{oursIG}{rgb}{1.0, 0.13, 0.32}

\definecolor{colMAE}{rgb}{0.65, 0.4, 0.5}
\definecolor{colMAEINOneK}{rgb}{0.5058823529411764, 0.4470588235294118, 0.7019607843137254}
\definecolor{colMAEIG}{rgb}{0.8549019607843137, 0.5450980392156862, 0.7647058823529411}
\definecolor{colCE}{rgb}{0.4, 0.4, 0.6}
\definecolor{colMAECE}{rgb}{0.2, 0.6, 0.3}
\definecolor{colCLIP}{rgb}{0.7, 0.7, 0.0}
\definecolor{colMAECLIP}{rgb}{0.5, 0.0, 0.0}

\newcolumntype{i}{>{\columncolor{Image}}c}
\newcolumntype{v}{>{\columncolor{Video}}c}
\newcolumntype{t}{>{\columncolor{ThreeD}}c}
\newcolumntype{I}{>{\columncolor{ImageLight}}c}
\newcolumntype{V}{>{\columncolor{VideoLight}}c}
\newcolumntype{T}{>{\columncolor{ThreeDLight}}c}

\newcommand{\fixme}[1]{{\color{red} \textbf{#1}}}

\makeatletter
\newcommand\thefontsize{The current font size is: \f@size pt}
\makeatother

\newcommand{\prept}{pre-pretraining\xspace}
\newcommand{\Prept}{Pre-pretraining\xspace}
\newcommand{\preptShort}{pre-pretrain\xspace}
\newcommand{\preptemph}{\emph{pre-}pretraining\xspace}

\newcommand{\lit}{LiT\xspace}
\newcommand{\mae}{MAE\xspace}

\newcommand{\swag}{SWAG\xspace}
\newcommand{\ce}{WSP\xspace}

\newcommand{\ours}{MAE$\rightarrow$WSP\xspace}
\newcommand{\oursShort}{MAWS\xspace}

\newcommand{\clip}{CLIP\xspace}

\newcommand{\openclip}{OpenCLIP\xspace}

\newcommand{\scaleViT}{Scale-ViT\xspace}

\newcommand{\dino}{DINO\xspace}
\newcommand{\florence}{Florence\xspace}
\newcommand{\msn}{MSN\xspace}
\newcommand{\xlmr}{XLM-R\xspace}
\newcommand{\vitDet}{ViTDet\xspace}
\newcommand{\dinovTwo}{DINOv2\xspace}

\newcommand{\swinL}{Swin-L\xspace}
\newcommand{\vit}{ViT\xspace}
\newcommand{\vitB}{ViT-B\xspace}
\newcommand{\vitL}{ViT-L\xspace}
\newcommand{\vitH}{ViT-H\xspace}
\newcommand{\vitg}{ViT-g\xspace}
\newcommand{\vitG}{ViT-G\xspace}
\newcommand{\vitTwoB}{ViT-2B\xspace}
\newcommand{\vitSixB}{ViT-6.5B\xspace}

\newcommand{\CocaModel}{CoCa-2B\xspace}
\newcommand{\SwinVTwoModel}{SwinV2-G\xspace}
\newcommand{\florenceModel}{CoSwin-H\xspace}

\newcommand{\igShort}{IG\xspace}
\newcommand{\igSizeShort}{IG-3B\xspace}
\newcommand{\igSize}{Instagram-3B\xspace}
\newcommand{\ig}{Instagram\xspace}

\newcommand{\inetOneK}{ImageNet-1k\xspace}
\newcommand{\inetOneKShort}{IN1k\xspace}
\newcommand{\inetFull}{ImageNet-21k\xspace}
\newcommand{\inetFullShort}{IN21k\xspace}
\newcommand{\inetReal}{ImageNet-ReaL\xspace}
\newcommand{\inetRealShort}{IN-ReaL\xspace}

\newcommand{\objectNet}{ObjectNet\xspace}
\newcommand{\objectNetShort}{ON\xspace}
\newcommand{\inetvTwo}{ImageNetv2\xspace}
\newcommand{\inetvTwoShort}{INv2\xspace}
\newcommand{\kinetics}{Kinetics-400\xspace}
\newcommand{\kineticsShort}{K400\xspace}

\newcommand{\petsShort}{Pets\xspace}
\newcommand{\inat}{iNaturalist-18\xspace}
\newcommand{\inatShort}{iNat18\xspace}

\newcommand{\sthsth}{Something Something-v2\xspace}
\newcommand{\sthsthShort}{SSv2\xspace}

\newcommand{\cocoShort}{COCO\xspace}
\newcommand{\lvisShort}{LVIS\xspace}

\newcommand{\jftThreeHund}{JFT-300M\xspace}
\newcommand{\jftThreeB}{JFT-3B\xspace}

\newcommand{\alignThreeSixB}{ALIGN-3.6B\xspace}
\newcommand{\alignOriginal}{ALIGN-1.8B\xspace}
\newcommand{\wit}{WIT-400M\xspace}

\newcommand{\laionTwoB}{LAION-2B\xspace}

\newcommand{\food}{Food-101\xspace}
\newcommand{\foodShort}{F-101\xspace}
\newcommand{\pmd}{PMD\xspace}

\newcommand{\objectDetshort}{O365\xspace}
\newcommand{\swinVTwoDataset}{IN-ext-70M\xspace}

\newcommand{\florenceDataset}{FLD-900M\xspace}
\newcommand{\objectsThreeSixFive}{Objects365\xspace}

\newcommand{\app}{\raise.17ex\hbox{$\scriptstyle\sim$}}

\newcommand{\sota}{state-of-the-art\xspace}

\newcommand{\pptrain}{pre-pretrain\xspace}
\newcommand{\Pptrain}{Pre-pretrain\xspace}
\newcommand{\pptraining}{pre-pretraining\xspace}

\iccvfinalcopy %

\def\iccvPaperID{20} %

\ificcvfinal\fi

\begin{document}

\title{The effectiveness of MAE \preptemph for billion-scale pretraining}

\author{
  {\normalsize Mannat Singh$^{*,\dagger}$ \quad Quentin Duval$^{*}$ \quad Kalyan Vasudev Alwala$^{*}$ \quad Haoqi Fan} \\
  {\normalsize Vaibhav Aggarwal \quad Aaron Adcock \quad Armand Joulin \quad Piotr Doll\'ar }\\
  {\normalsize Christoph Feichtenhofer \quad Ross Girshick \quad Rohit Girdhar \quad Ishan Misra} \\
  {\normalsize Meta AI} \\
  {\normalsize \url{https://github.com/facebookresearch/maws}}
}

\maketitle

\makeatletter{\renewcommand*{\@makefnmark}{}
\footnotetext{
  $^*$Equal technical contribution.
  $^\dagger$Project Lead.
}
\makeatother}

\ificcvfinal\thispagestyle{empty}\fi

\begin{abstract}

This paper revisits the standard pretrain-then-finetune paradigm used in computer vision for visual recognition tasks.
Typically, state-of-the-art foundation models are pretrained using large scale (weakly) supervised datasets with billions of images.
We introduce an additional pre-pretraining stage that is simple and uses the self-supervised \mae technique to initialize the model.
While \mae has only been shown to scale with the size of models, we find that it scales with the size of the training dataset as well.
Thus, our \mae-based pre-pretraining scales with both model and data size making it applicable for training foundation models.
Pre-pretraining consistently improves both the model convergence and the downstream transfer performance across a range of model scales (millions to billions of parameters), and dataset sizes (millions to billions of images).
We measure the effectiveness of pre-pretraining on 10 different visual recognition tasks spanning image classification, video recognition, object detection, low-shot classification and zero-shot recognition.
Our largest model achieves new state-of-the-art results on \inat (91.7\%), \inetReal (91.1\%), 1-shot \inetOneK (63.6\%), and zero-shot transfer on 
\food (96.2\%).
Our study reveals that model initialization plays a significant role, even for web-scale pretraining with billions of images, and our models are available publicly.
\end{abstract}

\section{Introduction}
\label{sec:intro}

\begin{figure}
  \centering
  \includegraphics[width=\linewidth]{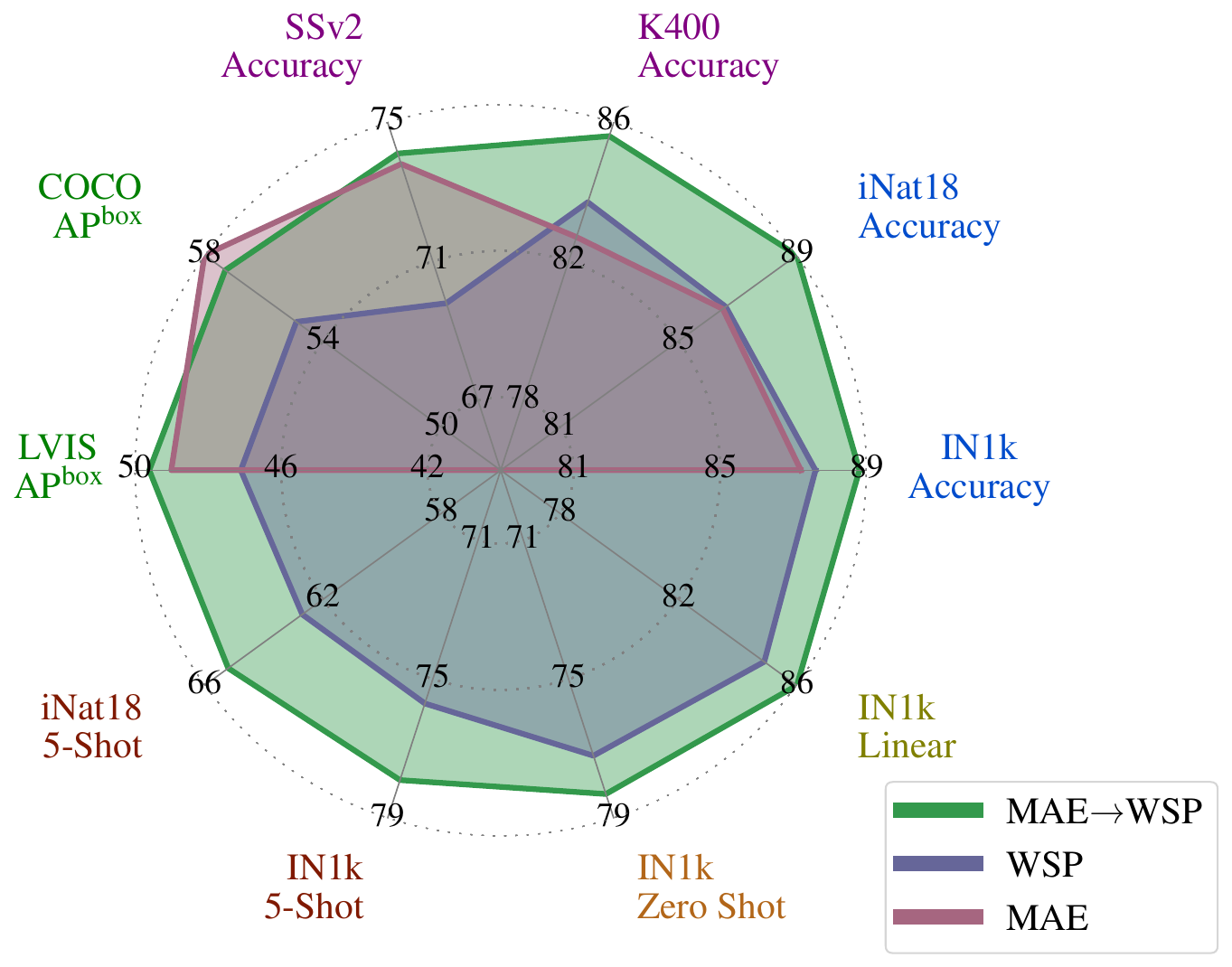}
    \ifarxiv
        \vspace{-0.2in}
    \else
        \vspace{-0.2in}
    \fi
  \caption{\textbf{\mae \prept improves performance}.
  Transfer performance of a \vitL architecture trained with self-supervised  pretraining (\mae), weakly supervised pretraining on billions of images (\ce), and our \prept (\ours) that initializes the model with \mae and then pretrains with \ce.
  \Prept consistently improves performance.
  }
  \label{fig:mae_vs_ce_vs_mae_ce_ig}
\end{figure}

The pretrain-then-finetune paradigm in visual recognition has enabled high performance visual recognition models across a range of tasks such as image classification~\cite{singh2022revisiting,mahajan2018exploring,radford2021learning}, video action recognition~\cite{girdhar2023omnimae,feichtenhofer2022masked,ghadiyaram2019large}, object detection~\cite{caron2020unsupervised,zhou2021ibot}, 3D \etc.
Typically, pretraining consists of training a model using a pretraining task on large scale data.
The resulting pretrained models learn general purpose visual representations that can be used for a range of target tasks, often with limited labeled data, by transfer learning.

In this paper, we show that an initial stage of \prept before the standard pretraining task can improve vision models across a variety of different tasks.
Our method combines two common pretraining tasks in vision: (1) weakly supervised pretraining that uses weak, often noisy, signals such as text or image hashtags as supervision, and (2) self-supervised pretraining that only uses the data without additional supervision.
Both forms of pretraining start training with a randomly initialized model and have proven effective at learning general purpose vision models.
While there have been attempts to combine both these forms of pretraining~\cite{mu2022slip, singh2022flava}, they are typically used independently in the pretrain-then-finetune two stage paradigm~\cite{dosovitskiy2020image, zhai2022scaling,singh2022revisiting}.

In this work we explore the combination of self- and weakly-supervised learning in a simple \textit{pre}-pretraining framework, as follows. 
We first begin with the Masked Autoencoder (\mae)~\cite{he2021masked} self-supervised learning technique to \pptrain vision models without using any labels.
After initializing from the pre-pretrained model, we use standard weakly supervised pretraining on billions of images with noisy labels.
We perform a large-scale empirical study to measure the effectiveness of pre-pretraining on 10 different visual recognition tasks spanning image classification, video recognition, object detection, low-shot classification and zero-shot recognition. 
Our study reveals that \pptrain initialization improves the performance for the weakly supervised models, and this improvement holds even at billion scale weakly labeled data, and across vision tasks (\cref{fig:mae_vs_ce_vs_mae_ce_ig}).
It also improves the model convergence during pretraining, leading to an efficient way of training large scale vision models.
\Pptrain further enjoys the computational efficiency of the MAE approach, making it simple and scalable.
Finally, we show that by using \pptraining, \emph{both} self-supervised learning and weakly supervised learning can be combined for improved model performance for billion-scale data.

\Pptrain is related to `intermediate finetuning'~\cite{bao2021beit, liu2022swin} which introduces a stage \emph{after} pretraining to better align the pretrained features with the downstream task using labeled data.
In contrast, \pptrain serves as a better way to initialize a model \emph{before} pretraining.
Since we leverage MAE for \pptraining, we do not need additional information or labels for this stage and can re-use the pretraining data.
This makes \pptrain convenient and simple to use with existing pretraining datasets.

Our study on large-scale \pptraining reveals that model initialization plays a significant role, even for web-scale pretraining, and \prept is a simple and promising technique in that direction. In particular, we show that
 (i) \mae not only scales with model size as shown in~\cite{he2021masked}, but \textit{also} with the size of the training \textit{data} (\cref{fig:mae_scaling}).
(ii) \Prept improves both the model convergence and the final downstream performance for different sized models (millions to billions of parameters) trained on different sized datasets (millions to billions of images).
(iii) Using \prept combines the benefits of both self-supervised learning and large scale weakly-supervised learning, and our models achieve excellent performance on a variety of different visual recognition tasks (\cref{fig:mae_vs_ce_vs_mae_ce_ig}). 
Most prominently, our model sets new state-of-the-art results on image classification on \inat (91.7\%) and \inetReal (91.1\%), 1-shot \inetOneK classification (63.6\%),
and zero-shot transfer on \food (96.2\%).

\section{Related Work}
\label{sec:related}

{\noindent \bf Supervised pretraining} of transferrable representations on large labeled datasets~\cite{deng2009imagenet,kay2017kinetics,ridnik2021imagenet} and employing them for downstream recognition tasks, has emerged as a powerful approach in computer vision.
It has spurred rapid progress on various tasks including image classification~\cite{donahue2014decaf,esc,razavian2014features}, object detection/segmentation~\cite{girshick2013rcnn,ren2015faster}, image captioning~\cite{li2020oscar,yu2022coca} and video action recognition~\cite{simonyan2014two,carreira2017quo,fan2021multiscale}.
While useful, such representations are often limited by the scale and diversity of the supervision in the pretraining datasets. Hence,
recent work has probed the effectiveness, robustness, and fairness of these representations
~\cite{abnar2021exploring,kornblith2019better,de2019does,recht2019imagenet,shankar2021image,taori2020measuring,idrissi2022imagenet}.

{\noindent \bf Self-supervised pretraining} is a promising alternative to learn these representation without relying on large well-labeled datasets.
Initial works focused on reconstructions methods~\cite{vincent2010stacked} before moving to other pretraining tasks
such as solving jigsaw puzzles~\cite{noroozi2016unsupervised},
constrastive learning~\cite{he2019moco,chen2020simple} or joint embedding
approaches~\cite{grill2020bootstrap,caron2020unsupervised,caron2021emerging, zhou2021ibot,assran2022masked,assran2023self}. %
With the advent of Vision Transformers \cite{dosovitskiy2020image}, approaches based on reconstructions such as \cite{bao2021beit,xie2021simmim,he2021masked} got renewed interest for their simplicity and state of the art performance.
Of particular interest to us is MAE~\cite{he2021masked} for its state of the art performance on many transfer tasks~\cite{he2021masked,li2022exploring,girdhar2023omnimae,tong2022videomae,feichtenhofer2022masked} and its computational efficiency.
Given the lack of supervision during pretraining, these representations often require significant finetuning to align to downstream tasks.

{\bf \noindent Weakly supervised pretraining (\ce)} is a middle-ground between supervised and self-supervised pretraining. Instead of ignoring annotations completely as in self-supervised pretraining, or requiring exhaustive labels as in supervised pretraining, \ce
relies on the large quantity of ``free'' annotation available on the internet. These annotations occur as image-text
pairs~\cite{radford2021learning,schuhmann2022laion}, where the text can additionally be processed to produce pseudo labels. %
Of particular interest to us is the latter, \ie
approaches which leverage multi-label classification on noisy labels~\cite{zhai2022scaling,mahajan2018exploring,singh2022revisiting,ghadiyaram2019large}
which have shown state of the art fine-tuning performance, and at the same time can be adapted using image-text data
to gain zero-shot capabilities~\cite{zhai2022lit}.
In this work, we explore \ce in conjunction with self-supervised \prept, and show faster convergence and stronger performance.

\section{Setup}

Our goal is to empirically study the effectiveness of self-supervised pre-pretraining as a precursor to billion scale weakly supervised pretraining for representation learning.
Given the simplicity and efficiency of Masked AutoEncoding (MAE)~\cite{he2021masked},
we leverage it as the self-supervised pre-pretraining approach. %
Our study shows that MAE scales with the size of the pretraining dataset and model size, and combining it with weak supervision improves large scale vision models.
Additionally, such a combination leads to faster convergence and is a simple, scalable way to learn visual representations at scale.
We describe our setup and the approaches in detail next.

\par \noindent \textbf{Architecure.}
We use the Vision Transformer (\vit)~\cite{dosovitskiy2020image} architecture as the visual encoder for all our experiments.
ViTs employ minimal vision-specific inductive biases combined with the standard transformer architecture~\cite{vaswani2017attention}, and yet have emerged as an architecture of choice for a wide variety of visual and multimodal recognition tasks~\cite{zhai2022scaling,arnab2021vivit,girdhar2023omnimae}.
We train \vit models
at various scales in terms of number of parameters, including \vitB (86M), \vitL (307M), and \vitH (632M). We also train on larger 1.9B and 6.5B parameter \vit models, which
we call \vitTwoB and \vitSixB, respectively (\cref{tab:app_model_architectures}).
As is common practice \cite{dosovitskiy2020image,zhai2022scaling}, we train models of sizes \vitB, \vitL with a patch size
of 16 and larger models with a patch size of 14. We pretrain with a $224\times{}224$ resolution for all models. 

\begin{table}[!htb]
    \centering
    \resizebox{0.7\linewidth}{!}{
    \setlength{\tabcolsep}{2pt}
    \begin{tabular}{l|cccc|c}
        \bf Arch.  & \bf Layers & \bf Embed & \bf MLP & \bf Heads & \bf Params \\
        \midrule
        \vitB   & 12 & 768 & 3072 & 12 & 86M \\
        \vitL   & 24 & 1024 & 4096 & 16 & 307M \\
        \vitH   & 32 & 1280 & 5120 & 16 & 632M \\
        \vitTwoB & 24 & 2560 & 10240 & 32 & 1.89B \\
        \vitSixB & 32 & 4096 & 16384 & 32 & 6.44B \\
    \end{tabular}
    }
    \caption{
        \textbf{Model architecture details.} To ensure we were able to scale models out further than \vitH easily, we decided to scale 
        models along the same lines as GPT-3~\cite{brown2020language}, which has proven to be successful for NLP.
    }
    \label{tab:app_model_architectures}
\end{table}

\par \noindent \textbf{Pre-pretraining (\mae)}~\cite{he2021masked} learns visual representations from image datasets without using any labels. %
We choose this approach as it is simple to implement and scales very effectively with large \vit model sizes due to patch dropping as described next. %
\mae randomly masks 75\% of an image %
and trains the model to reconstruct the masked input image by minimizing the pixel reconstruction error.
The target pixel values for a given patch are normalized by the mean and standard deviation of all pixels in it.
Coupled with the \vit architecture, \mae can be trained by only processing the 25\% unmasked image patches.
A separate, smaller, decoder is then used to reconstruct the missing part of the input.
This asymmetrical design makes training the encoder extremely efficient, allowing for scaling visual encoder sizes.

\par \noindent \textbf{Weakly-supervised pretraining (\ce)} leverages images with associated `weak' supervision for training models.
In particular, we focus on internet images and use their associated text information as supervision.
We convert the text into a discrete set of labels, specifically leveraging hash-tag information~\cite{mahajan2018exploring,singh2022revisiting,ghadiyaram2019large}.
We then use a multi-label classification loss to train models.
We refer to this method as \ce. %

\par \noindent \textbf{\ours}, or \textbf{\oursShort} for short,
first trains the encoder %
using the \mae self-supervised method using only the images. %
This pre-pretraining stage initializes the model while simultaneously being computationally efficient because of the masking used in \mae.
In the second stage, we pretrain the encoder using both the image and associated weak supervision. %
This combination outperforms using either strategy in isolation, \ie, an \mae model or a weakly supervised model trained from scratch.

\section{Experiments}

We empirically evaluate and analyze large scale \mae \prept using \ig data on a variety of different visual recognition tasks.
We describe the datasets used for pretraining and evaluation, followed by analysis of 
the pretraining design decisions, and finally the downstream transfer evaluation of our learned representation.

\begin{figure*}[!t]
\begin{center}
  \captionsetup{type=figure}
    \begin{subfigure}{.245\textwidth}
        \begin{tikzpicture}
    \begin{axis}[
        legend pos=south east,
        xmin=0,
        xmode=log,
        grid=both,
        grid style={line width=.1pt, draw=gray!10},
        major grid style={line width=.2pt,draw=gray!50},
        minor tick num=2,
        log ticks with fixed point,
        xtick={0.1, 0.5, 2.0, 6.0},
        axis x line*=bottom,
        axis y line*=left,
        height=1.4\linewidth,
        width=1.2\linewidth,
        ylabel style= {align=center, font=\small},
        title style= {align=center, font=\small},
        xlabel style = {font=\small},
        title={{\color{ImageDark}\inetOneKShort (accuracy)}},
        xlabel={Model parameters (billions)},
        yticklabel style = {font=\small},
        xticklabel style = {font=\small},
        legend style={cells={align=left}, font=\footnotesize, fill=none, draw=none},
    ]
    \addplot[mark=*, colMAE, very thick, mark options={solid}] plot coordinates {
        (0.086, 83.5)  %
        (0.307, 86.1) %
        (0.632, 87.4) %
        (1.9,  87.8) %
        (6.5,  88.3)
    };
    \addlegendentry{\igSizeShort}
    \addplot[mark=o, colMAE, very thick, densely dashed, mark options={solid}] plot coordinates {
        (0.086, 83.5)  %
        (0.307, 86) %
        (0.632, 86.9) %
        (1.9,  87.4) %
    };
    \addlegendentry{\inetOneKShort}
    \end{axis}
\end{tikzpicture}
    \end{subfigure} \hfill
    \begin{subfigure}{.245\textwidth}
        \begin{tikzpicture}
    \begin{axis}[
        legend pos=south east,
        xmin=0,
        xmode=log,
        grid=both,
        grid style={line width=.1pt, draw=gray!10},
        major grid style={line width=.2pt,draw=gray!50},
        minor tick num=2,
        log ticks with fixed point,
        xtick={0.1, 0.5, 2.0, 6.0},
        axis x line*=bottom,
        axis y line*=left,
        height=1.4\linewidth,
        width=1.2\linewidth,
        title style= {align=center, font=\small},
        ylabel style= {align=center, font=\small},
        xlabel style = {font=\small},
        title={{\color{ImageDark}\inatShort (accuracy)}},
        xlabel={Model parameters (billions)},
        yticklabel style = {font=\small},
        xticklabel style = {font=\small},
        legend style={cells={align=left}, font=\footnotesize, fill=none, draw=none},
    ]
    \addplot[mark=*, colMAE, very thick, mark options={solid}] plot coordinates {
        (0.086, 74.7) %
        (0.307, 80.7) %
        (0.632, 84.0) %
        (1.9, 85.6) %
        (6.5, 86.6)
    };
    \addlegendentry{\igSizeShort}
    \addplot[mark=o, colMAE, very thick, densely dashed, mark options={solid}] plot coordinates {
        (0.086, 75.0)  %
        (0.307, 80.2)  %
        (0.632, 82.8) %
        (1.9,  84.5) %
    };
    \addlegendentry{\inetOneKShort}
    \end{axis}
\end{tikzpicture}
    \end{subfigure} \hfill
    \begin{subfigure}{.245\textwidth}
        \begin{tikzpicture}
    \begin{axis}[
        legend pos=south east,
        xmin=0,
        xmode=log,
        grid=both,
        grid style={line width=.1pt, draw=gray!10},
        major grid style={line width=.2pt,draw=gray!50},
        minor tick num=2,
        log ticks with fixed point,
        xtick={0.1, 0.5, 2.0},
        ytick={44, 49, 54},
        axis x line*=bottom,
        axis y line*=left,
        height=1.4\linewidth,
        width=1.2\linewidth,
        title style= {align=center, font=\small},
        ylabel style= {align=center, font=\small},
        xlabel style = {font=\small},
        title={{\color{DetDark}\lvisShort (AP\textsuperscript{box})}},
        xlabel={Model parameters (billions)},
        yticklabel style = {font=\small},
        xticklabel style = {font=\small},
        legend style={cells={align=left}, font=\footnotesize, fill=none, draw=none},
    ]
    \addplot[mark=*, colMAE, very thick, mark options={solid}] plot coordinates {
        (0.086, 42.9)
        (0.307, 49.0)
        (0.632, 52.7)
        (1.9, 53.6)
    };
    \addlegendentry{\igSizeShort}
    \addplot[mark=o, colMAE, very thick, densely dashed, mark options={solid}] plot coordinates {
        (0.086, 43.0)
        (0.307, 49.2)
        (0.632, 51.5)
    };
    \addlegendentry{\inetOneKShort}
    \end{axis}
\end{tikzpicture}
    \end{subfigure} \hfill
    \begin{subfigure}{.245\textwidth}
        \begin{tikzpicture}
    \begin{axis}[
        legend pos=south east,
        xmin=0,
        xmode=log,
        grid=both,
        grid style={line width=.1pt, draw=gray!10},
        major grid style={line width=.2pt,draw=gray!50},
        minor tick num=2,
        log ticks with fixed point,
        xtick={0.1, 0.5, 2.0},
        ytick={54, 57, 60},
        axis x line*=bottom,
        axis y line*=left,
        height=1.4\linewidth,
        width=1.2\linewidth,
        title style= {align=center, font=\small},
        ylabel style= {align=center, font=\small},
        xlabel style = {font=\small},
        title={{\color{DetDark}\cocoShort (AP\textsuperscript{box})}},
        xlabel={Model parameters (billions)},
        yticklabel style = {font=\small},
        xticklabel style = {font=\small},
        legend style={cells={align=left}, font=\footnotesize, fill=none, draw=none},
    ]
    \addplot[mark=*, colMAE, very thick, mark options={solid}] plot coordinates {
        (0.086, 53.8)
        (0.307, 58.0)
        (0.632, 59.1)
        (1.9, 59.9)
    };
    \addlegendentry{\igSizeShort}
    \addplot[mark=o, colMAE, very thick, densely dashed, mark options={solid}] plot coordinates {
        (0.086, 54.0)
        (0.307, 57.6)
        (0.632, 58.7)
    };
    \addlegendentry{\inetOneKShort}
    \end{axis}
\end{tikzpicture}
    \end{subfigure}
    \ifarxiv
        \vspace{-0.23in}
    \else
        \vspace{-0.4in}
    \fi
    \caption{\textbf{Scaling MAE with model and dataset size}. We plot \mae's performance when pretrained on \inetOneK or \igSize and
    finetuned on downstream tasks. \mae scales to billion parameters sized models using just \inetOneKShort pretraining.
    Larger models show improved scaling behavior when pretrained with the much larger \igSizeShort dataset.
    Tabulated results in Appendix \cref{tab:mae_scaling_numbers}.
    \inetOneKShort and \inatShort results are finetuned at 224px resolution.
    For \cocoShort and \lvisShort, \mae pretrained on \inetOneKShort for \vitTwoB is missing as training at that scale was unstable, and \vitSixB results are skipped due to compute limitations.
    }
    \label{fig:mae_scaling}
\end{center}
\end{figure*}

\subsection{Datasets and training details}

\begin{table}
    \centering
    \resizebox{\linewidth}{!}{
    \setlength{\tabcolsep}{2pt}
    \begin{tabular}{l|cccc}
        \bf Dataset  & \bf Task & \bf \#cls & \bf \#train & \bf \#val \\
        \midrule
        {\color{ImageDark}{\inetOneK}} (\inetOneKShort)~\cite{ILSVRC15} & Image cls. & 1000 & 1M & 50K \\
        {\color{ImageDark}{\inat}} (\inatShort)~\cite{iNaturalist} & Fine-grained cls. & 8142 & 437K & 24K \\
        {\color{ImageDark}{\inetvTwo}} (\inetvTwoShort)~\cite{recht2019imagenet} & Image cls. & 1000 & -- & 10K \\
        {\color{ImageDark}{\inetReal}} (\inetRealShort)~\cite{beyer2020we} & Image cls. & 1000 & -- & 50K \\
        {\color{ImageDark}{\objectNet}} (\objectNetShort)~\cite{barbu2019objectnet} & Image cls. & 113 & -- & 19K \\
        {\color{ImageDark}{\food}} (\foodShort)~\cite{bossard2014food} & Image cls. & 101 & N\slash{}A & 25K \\
        {\color{DetDark}{\cocoShort}}~\cite{lin2014microsoft} & Obj.\ det. & 80 & 118K & 5K \\
        {\color{DetDark}{\lvisShort}}~\cite{gupta2019lvis} & Obj.\ det. & 1K & 100K & 20K \\
        {\color{VideoDark}{\kinetics}} (\kineticsShort)~\cite{kay2017kinetics} & Action cls. & 400 & 220K & 20K \\
        {\color{VideoDark}{\sthsth}} (\sthsthShort)~\cite{goyal2017something} & Action cls. & 174 & 169K & 25K \\
    \end{tabular}
    }
    \caption{\textbf{Evaluation datasets} used to evaluate \ours on {\color{ImageDark} image classification}, {\color{DetDark} object detection}, and {\color{VideoDark} video action recognition} tasks.
    The table reports the task, number of classes (\#cls), number of training samples (\#train), and number of validation samples (\#val) for each dataset.
    }
    \label{tab:eval_datasets}
\end{table}

\noindent \textbf{Pretraining dataset.} We use \igSize(\igSizeShort) a billion-scale multi-label dataset
sourced from \ig (\igShort). 
This multi-label dataset contains 28K classes and 3B unique images, resampled to 5B total images,
and was produced by running the dataset generation pipeline from \swag~\cite{singh2022revisiting} without modification.
Compared to~\cite{singh2022revisiting}, our
version of the dataset has 16\% fewer images (3.0B \vs 3.6B), but we were able to reproduce the results
from~\cite{singh2022revisiting} with our version.
We obtain labels using an automated process wherein we first obtain hashtags from the associated image captions, and
then map the hashtags to WordNet synsets following~\cite{singh2022revisiting}.
After this processing, we get the weakly labeled \igShort dataset that contains images and their associated labels.

\par \noindent \textbf{Evaluation datasets.}
We evaluate \ours on a variety of different downstream visual recognition tasks.
To evaluate our model on image classification, we use the standard \inetOneK~\cite{ILSVRC15} (\inetOneKShort) dataset,
and also the long-tailed and fine-grained \inat~\cite{iNaturalist} (\inatShort) dataset.
For object detection and segmentation, we use the popular \cocoShort~\cite{lin2014microsoft} dataset, and also \lvisShort
\cite{gupta2019lvis}, a large vocabulary dataset for long tailed object recognition.
We evaluate video classification performance using two popular action recognition datasets, \kinetics~\cite{kay2017kinetics}
(\kineticsShort) and \sthsth~\cite{goyal2017something} (\sthsthShort). For zero-shot transfer, we evaluate on \inetOneKShort
and \food~\cite{bossard2014food} (\foodShort).
We also evaluate the robustness of our models on test sets which overlap with \inetOneKShort classes, specifically
\inetvTwo~\cite{recht2019imagenet} (\inetvTwoShort), \inetReal~\cite{beyer2020we} (\inetRealShort),
and \objectNet~\cite{barbu2019objectnet} (\objectNetShort). 
Please see~\cref{tab:eval_datasets} for more details.

\par \noindent \textbf{\mae pretraining details.}
We follow \cite{he2021masked} to train \mae models on \igSizeShort without using any labels.
We mask 75\% of the image for this training and train the model for 1 epoch over the dataset. We follow the same hyperparameters
used in \cite{he2021masked} for pretraining on \inetOneKShort.

\par \noindent \textbf{Supervised pretraining details.}
We train with a supervised cross-entropy loss on \igSizeShort using the hashtags as labels.
This model is trained by default with random weight initialization and we use the training hyperparameters from~\cite{singh2022revisiting}.
\par \noindent \textbf{Using \prept.}
When using \prept, we first train a model from scratch using \mae on the \igShort dataset.
We then use the weights of the \mae encoder and perform supervised pretraining using the cross-entropy loss as described above.
We reuse the same hyperparameters and training details as
\cite{singh2022revisiting}, \ie \emph{there is no hyperparameter search needed} for \ours, and we train for 1 epoch on
\igSizeShort.

\par \noindent \textbf{Zero-shot training and evaluation details.}
To impart zero shot understanding capabilities to our models, we use the \lit approach from~\cite{zhai2022lit}.
For \lit, we use the original \emph{(image, caption)} pairs from the \igSizeShort dataset.
We \emph{freeze} the image encoder, and train a text encoder to encode the image captions and match the text embeddings to the associated image embedding using a \clip loss~\cite{radford2021learning}.
We train the text encoder for 1 epoch.
For evaluation, we follow~\cite{radford2021learning} -- we use the text encoder to compute embeddings from the templated text descriptions of classes and use the cosine similarity of the image and text embeddings as the classification score.

For full training details and hyperparameters, please refer to \cref{app:pretraining_details}.

\begin{figure*}
    \centering
    \begin{minipage}[t]{.29\textwidth}
        \centering
        \input{figures/mae_for_wsl_model_scaling.tex}
    \end{minipage}\hfill
    \begin{minipage}[t]{0.39\textwidth}
        \centering
        \input{figures/mae_vs_ce_epochs.tex}
    \end{minipage}\hfill
    \begin{minipage}[t]{0.29\textwidth}
        \centering
        \input{figures/mae_for_wsl_flops_scaling.tex}
    \end{minipage}
\end{figure*}

\subsection{Scaling \mae pretraining to large data}
\label{sec:mae-scale}

Since our \prept uses \mae in the very first stage, we first study how \mae behaves on the large scale \igSizeShort dataset.
We compare the performance of \mae pretraining on the large scale \igSizeShort with the original \mae~\cite{he2021masked} models 
trained on \inetOneKShort for 1600 epochs.
We train models of varying sizes, from \vitB to \vitH as in~\cite{he2021masked}.
To test the scaling behavior further, we also train \mae on \vitTwoB and \vitSixB, with 2B and 6.5B parameters, respectively.
We measure the performance of the resulting models in
\cref{fig:mae_scaling} on four different vision tasks.

We observe that using the \igSizeShort data provides consistent gains over \inetOneKShort for all vision tasks, and the gain increases for larger models.
These experiments show that \mae scales with the size of the pretraining dataset, and benefits from using billions of images from \igSizeShort.
He \etal~\cite{he2021masked}'s findings were limited to the fact that \mae scales with the size of the model, and thus our findings on \mae scaling with the size of the pretraining data are complementary to theirs.

Our \vitTwoB model pretrained on \igSizeShort improves upon the best results from \cite{he2021masked} on image classification,
attaining 87.8\% on \inetOneKShort (+0.9\%) and 85.6\% on \inatShort (+2.6\%) at 224$\times$224 resolution. 
The gains on detection are equally encouraging, with our \vitTwoB
reaching 53.6 AP\textsuperscript{box} on \lvisShort  (+2.1 over \cite{he2021masked}) and 59.8 AP\textsuperscript{box} on \cocoShort
(+1.2 over \cite{he2021masked}). Our \vitSixB model further improves the downstream performance,
attaining 88.3\% on \inetOneKShort and 86.6\% on \inatShort at just 224$\times$224 resolution.

Lastly, we highlight the simplicity of our setup, since we use the same hyperparameters as~\cite{he2021masked} to train \mae on 
\igSizeShort, even for the $3\times$ and $10\times$ larger \vitTwoB and \vitSixB models, without requiring extra tweaks. 
We note that
training the \vitSixB was sensitive to numerical stability issues, so we trained the model with full precision to avoid divergence.

\subsection{\mae \prept}
\label{sec:pretraining}

\begin{figure}
  \centering
  \includegraphics[width=\linewidth]{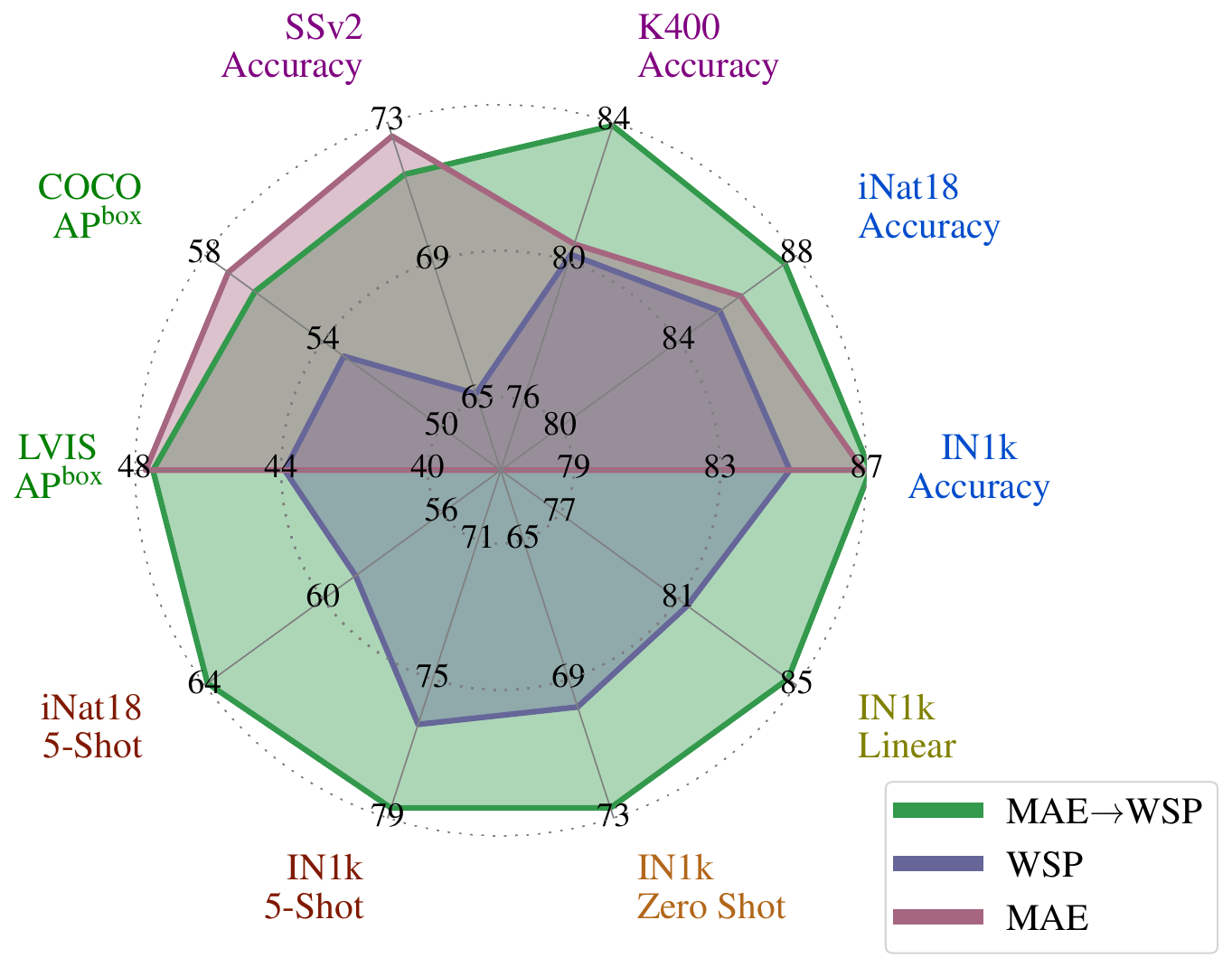}
    \ifarxiv
        \vspace{-0.2in}
    \else
        \vspace{-0.2in}
    \fi
  \caption{\textbf{MAE initialization for medium scale data}. Results for a \vitL trained on \inetFullShort with \mae, \ce, and \ours.
  \mae \prept improves \ce results by a wide margin. 
  }
  \label{fig:mae_vs_ce_vs_mae_ce_in21k}
\end{figure}

Given the promising aspects of \mae as a pretraining approach from~\cref{sec:mae-scale}, specifically that \mae (i) trains with larger models and datasets
without needing any tuning (ii) shows gains when scaling model and / or dataset size (iii) is efficient to train, we investigate it
as a \preptemph approach for supervised pretraining (\ce).

\cref{fig:mae_vs_ce_vs_mae_ce_ig} shows the performance of a \vitL with \mae pretraining, supervised pretraining (\ce), or
\mae \prept followed by supervised pretraining (\ours). We see that \mae and \ce have different strengths. \mae has strong performance for object detection, and full finetuned image classification. However, \mae underperforms on tasks where the model
is not finetuned, such as linear
classifiers, zero-shot, or low-shot classification -- situations where \ce performs better. For
these evaluations \mae lags behind \ce by more than 10 points, which is why the results for \mae are not visible in~\cref{fig:mae_vs_ce_vs_mae_ce_ig}.
For video classification, \mae performs significantly better than \ce on \sthsthShort, but lags behind it on \kineticsShort.

\ours outperforms either of \mae or \ce pretraining on most evaluations, across image classification, video recognition, zero-shot evaluation, object detection, \etc.
Given that all baselines are trained at billion-scale, these results show that \ours is a simple yet promising strategy to improve performance while requiring no extra data or tuning.
Next, we ablate the key aspects of \ours.

\begin{table*}
    \begin{minipage}{.57\linewidth}
        \begin{table}[H]
    \centering
    \setlength{\tabcolsep}{3pt}
    \resizebox{\linewidth}{!}{
    \begin{tabu}{llcc|cccc|c}
        \centering
        \multirow{2}{*}{\bf Method} & \multirow{2}{*}{\bf Dataset} & \multirow{2}{*}{\bf Arch.} & \multirow{2}{*}{\bf Res.} & \multicolumn{4}{c|}{\bf \inetOneK} & \multirow{2}{*}{\bf \inatShort} \\
        &&&& \bf \inetOneKShort & \bf \inetvTwoShort & \bf ReaL & \bf \objectNetShort \\
        \midrule
        \mae~\cite{he2021masked} & \inetOneKShort & \vitH & 448 & 87.8 & -- & -- & -- & 86.8 \\
        \swag~\cite{singh2022revisiting} & IG-3.6B & \vitH & 518 & 88.6 & 81.1 & 90.5 & 69.5 & 86.0 \\
        \dinovTwo~\cite{oquab2023dinov2} & LVD-142M & ViT-g & 448 & 88.9 & -- & -- & -- & -- \\
        \florence~\cite{yuan2021florence} & \florenceDataset & \florenceModel & 512 & 90.1 & -- & -- & -- & -- \\
        ViT~\cite{dosovitskiy2020image} & \jftThreeHund & \vitH & 518 & 88.6 & -- & 90.7 & -- & -- \\
        \scaleViT~\cite{zhai2022scaling} & \jftThreeB & \vitL & 384 & 88.5 & 80.4 & 90.4 & -- & -- \\
        \scaleViT~\cite{zhai2022scaling} & \jftThreeB & \vitG & 518 & 90.5 & 83.3 & 90.8 & 70.5 & -- \\
        SwinV2~\cite{liu2022swin} & \swinVTwoDataset & \SwinVTwoModel & 640 & 90.2 & \textbf{84.0} & -- & -- & -- \\
        CoCa~\cite{yu2022coca} & \makecell[l]{\jftThreeB + \\ \alignOriginal} & \CocaModel & 576 & \textbf{91.0} & -- & -- & -- & -- \\
        \midrule
        \oursShort & \igSizeShort & \vitH & 518 
        & 89.3 %
        & 82.3 %
        & 90.8 %
        & 72.6 %
        & 90.5 %
        \\
        \oursShort & \igSizeShort & \vitTwoB & 518 
        & 89.7 %
        & 83.0 %
        & 90.9 %
        & 75.8 %
        & 91.3 %
        \\ 
        \oursShort & \igSizeShort & \vitSixB & 518 
        & 90.1
        & \textbf{84.0}
        & \textbf{91.1}
        & \textbf{77.9}
        & \textbf{91.7}
        \\ 
    \end{tabu}
    }
    \caption{\textbf{Image classification results.} We report finetuning results on \inetOneKShort and \inatShort and note the pretraining dataset, architecture and finetuning resolution. 
    We also 
    evaluate the performance of our \inetOneKShort finetuned models on multiple test sets to measure robustness. 
    Our models are robust and also push the \sota on the challenging
    fine-grained and long-tailed \inatShort dataset.
    }
    \label{tab:image_classification}
\end{table}

    \end{minipage}
    \hspace{0.03in}
    \begin{minipage}{.43\linewidth}
        \begin{table}[H]
    \centering
    \setlength{\tabcolsep}{3pt}
    \resizebox{\linewidth}{!}{
    \begin{tabu}{llcc|c|c}
        \centering
        \bf Method & \bf Dataset & \bf Arch. & \bf Res. & \bf \kineticsShort & \bf \sthsthShort \\
        \midrule
        \florence~\cite{yuan2021florence} & \florenceDataset & \florenceModel & 384 & 86.5 & -- \\
        SwinV2~\cite{liu2022swin} & \swinVTwoDataset & \SwinVTwoModel & 384 & 86.8 & -- \\
        CoCa~\cite{yu2022coca} & \makecell[l]{\jftThreeB + \\ \alignOriginal} & \CocaModel & 576 & \textbf{88.9} & -- \\
        \midrule
        \multicolumn{6}{l}{\em Results with models pretrained on videos} \\
        MaskFeat~\cite{wei2022masked} & \kineticsShort & MViT-L & 224 & 84.3 & -- \\
        MAE~\cite{feichtenhofer2022masked,tong2022videomae} & \kineticsShort & \vitL & 224 & 85.2 & 74.0 \\
        OmniMAE~\cite{girdhar2023omnimae} & \inetOneKShort + \sthsthShort & \vitL & 224 & 84.0 & 74.2 \\
        MAE~\cite{feichtenhofer2022masked,tong2022videomae} & \kineticsShort & \vitH & 224 & 86.6 & -- \\
        \midrule
        \oursShort & \igSizeShort & \vitL & 224 
        & 86.0 %
        & \textbf{74.4} %
        \\
    \end{tabu}
    }
    \caption{\textbf{Video classification results} on \kinetics and \sthsth. Our models generalize well to video action 
    recognition tasks, despite not seeing any videos during pretraining. 
    }
    \label{tab:video_classification}
\end{table}

    \end{minipage}
\end{table*}

\par \noindent \textbf{Effect of model size.}
In~\cref{fig:mae_for_wsl_model_scaling} we study the effect of \prept compared to random initialization for different model sizes.
After initialization, all models are pretrained using \ce on the \igSizeShort dataset, and we measure the transfer performance on \inetOneKShort using linear probing.
We observe that \mae \prept gives consistent gains over the \ce baseline across all model sizes, ranging from 86M to 6.5B parameters.
The gains over the \ce baseline \emph{increase} for larger model sizes showing that \prept shows promising scaling behavior with model sizes.
Notably, a 2B \ours model outperforms a larger 6.5B \ce model.

\par \noindent \textbf{Number of \prept epochs.}
We vary the number of \mae \prept and the number of \ce pretraining epochs to understand their effect on the final recognition performance.
We study this in~\cref{fig:mae_vs_ce_epochs}.

\Prept improves results over the standard pretraining (random initialization, w/o \prept), and provides large gains with fewer \ce pretraining epochs.
\Prept also leads to faster convergence since even a small amount of \prept for 0.1 epochs provides improvements.
Increasing the epochs of \prept provide a larger improvement, and the gains saturate at 1 epoch of \prept.
Finally, \prept's gains do not diminish even after 4 epochs of \ce (20 billion samples) showing the value of \prept at scale.
We also note that these gains are independent of the evaluation protocol, and we observed them with full finetuning on \inetOneKShort as well.

\par \noindent \textbf{Training efficiency.}
\cref{fig:mae_for_wsl_flops_scaling} shows a comparison between \ce and \ours when comparing training FLOPs.
For the same training FLOPs, \ours achieves better transfer performance compared to \ce, and is up to $2\times$ more efficient.
\Prept's training efficiency holds over a large $10\times$ compute window.

\par \noindent \textbf{Different datasets for \prept.}
We also evaluate the performance of \ours when \prept \mae on the much smaller \inetOneK dataset below, and find that \prept remains just as effective. 
This allows reusing pretrained \mae models in practice.

\begin{center}
    \resizebox{0.7\linewidth}{!}{
        \begin{tabu}{ll|c|cc}
            \makecell[c]{\bf \mae \\ \bf Dataset} & \makecell[c]{\bf \ce \\ \bf Dataset} & \bf Arch. & \bf \inetOneKShort & \bf \inatShort \\
            \midrule
            \igSizeShort & \igSizeShort & \vitH 
            & 89.3 %
            & 90.5 %
            \\
            \inetOneKShort & \igSizeShort & \vitH 
            & 89.4 %
            & 90.5 %
            \\
        \end{tabu}
    }
\end{center}

\par \noindent \textbf{Different datasets for \prept and pretraining.}
We investigate the effect of the dataset used in \prept and pretraining by using \inetFullShort~\cite{deng2009imagenet} for all methods, including \mae and \ours.
Compared to \igSizeShort, \inetFullShort is more curated, smaller (14M images), and has cleaner labels (21K classes), where each image is labeled with one class from the WordNet synsets~\cite{miller1995wordnet}.
For evaluating
zero shot performance, we use the \pmd~\cite{singh2022flava} dataset for \lit training.
For full details about the hyperparameters, refer to \cref{app:pretraining_details}.

\cref{fig:mae_vs_ce_vs_mae_ce_in21k} compares the performance of \mae, \ce and \ours when pretrained on this \inetFullShort
dataset.
We notice a similar trend as when pretraining on \igSizeShort where \ours outperforms both \mae and \ce.
This shows that \mae \prept works with datasets of different scales and distributions.

\subsection{Transfer Evaluation}
\label{sec:results}

We compare with \sota research on image and video classification, detection and segmentation, low-shot image classification, zero shot 
transfer, robustness analysis. For brevity, we refer to \ours as \oursShort in this section.

\begin{table*}
    \begin{minipage}{.58\linewidth}
        \begin{table}[H]
    \centering
    \setlength{\tabcolsep}{3pt}
    \resizebox{\linewidth}{!}{
    \begin{tabu}{llcH|ccc|ccc}
        \centering
        \multirow{2}{*}{\bf Method} & \multirow{2}{*}{\bf Dataset} & \multirow{2}{*}{\bf Arch.} & \bf Protocol & \multicolumn{3}{c|}{\bf \inetOneKShort} & \multicolumn{3}{c}{\bf \inatShort} \\ 
        & & & & \bf 1-shot & \bf 5-shot & \bf 10-shot & \bf 1-shot & \bf 5-shot & \bf 10-shot \\
        \midrule
        \multicolumn{6}{l}{\em Results with different pretraining datasets} \\
        \clip \cite{radford2021learning} & \wit & \vitL & Linear & 41.3 & 66.2 & 71.3 & 21.9 & 49.0 & 58.5 \\
        \openclip \cite{ilharco_gabriel_2021_5143773} & \laionTwoB & \vitH & Adapter & 44.3 & 70.0 & 74.9 & 26.0 & 54.6 & 63.7 \\
        \openclip \cite{ilharco_gabriel_2021_5143773} & \laionTwoB & \vitG & Adapter & 46.3 & 72.9 & 77.2 & 26.3 & 55.7 & 65.1 \\
        \scaleViT\textsuperscript{\textdagger}\cite{zhai2022scaling} & \jftThreeB & \vitG & Linear & -- & \textbf{83.0} & \textbf{84.9} & -- & -- & -- \\
        \midrule
        \dino \cite{caron2021emerging} & \inetOneKShort & ViT-B/8 & Linear & 45.8 & 64.6 & 69.0 & 19.8 & 45.9 & 55.9 \\
        \msn \cite{assran2022masked} & \inetOneKShort & ViT-L/7 & Linear & 57.1 & 72.1 & 74.4 & 17.0 & 38.0 & 48.1 \\
        \mae\cite{he2021masked} & \inetOneKShort & \vitH & Finetune & -- & 57.9 & 70.8 & -- & 48.5 & 68.2 \\
        \swag\cite{singh2022revisiting} & IG-3.6B & \vitH & Adapter & 59.4 & 78.7 & 81.0 & 30.1 & 62.8 & 72.3 \\
        \midrule
        \oursShort & \igSizeShort & \vitH & Adapter & 57.1 & 79.8 & 82.5 & 31.7 & 67.8 & 76.1 \\
        \oursShort & \igSizeShort & \vitTwoB & Adapter & 62.1 & 81.5 & 83.7 & 35.5 & 72.8 & 80.3 \\
        \oursShort & \igSizeShort & \vitSixB & Adapter & \textbf{63.6} & 82.6 & 84.6 & %
        \textbf{36.4} & \textbf{73.7} & \textbf{80.9}  %
        \\ 
    \end{tabu}%
    }
    \caption{
        \textbf{Low shot image classification.} We compare the performance of our models using just a few examples per class for 
        \inetOneK and \inat. \oursShort excels at classification even with just 1 example per class. Pretraining on large scale data 
        can outperform techniques designed to work well in the low data regime, like \msn.
        We evaluate and report results for each technique with the best protocol, except for when the checkpoints are not 
        available\textsuperscript{\textdagger}.
    }
    \label{tab:low_shot_image_classification}
\end{table}

    \end{minipage}
    \hspace{0.03in}
    \begin{minipage}{.42\linewidth}
        \begin{table}[H]
    \centering
    \setlength{\tabcolsep}{3pt}
    \resizebox{\linewidth}{!}{
    \begin{tabu}{llcHc|c|cH}
        \centering
        \bf Method & \bf Dataset & \bf Arch. & \bf Text Enc. & \bf Res. & \bf \inetOneKShort & \bf \foodShort & \bf \petsShort \\
        \midrule
        \multicolumn{6}{l}{\em Results with different pretraining datasets} \\
        \clip~\cite{radford2021learning} & \wit & ViT-L/14 & & 336 & 76.2 & 93.8 & 93.5  \\
        \openclip~\cite{ilharco_gabriel_2021_5143773} & \laionTwoB & \vitH & & 224 & 78.0 & 92.5 & 94.4 \\
        \openclip~\cite{ilharco_gabriel_2021_5143773} & \laionTwoB & \vitG & & 224 & 80.1 & 92.9 & 95.2 \\
        \florence~\cite{yuan2021florence} & \florenceDataset & \florenceModel &  & 384 & 83.7 & 95.1 & 95.9 \\
        \makecell[l]{\scaleViT~\cite{zhai2022scaling} \\ + \lit~\cite{zhai2022lit}} & \makecell[l]{\jftThreeB \\ + \alignThreeSixB} & \vitL & \vitH & 224 & 80.8 & -- & -- \\
        \makecell[l]{\scaleViT~\cite{zhai2022scaling} \\ + \lit~\cite{zhai2022lit}} & \makecell[l]{\jftThreeB \\ + \alignThreeSixB} & \vitg & \vitg & 288 & 85.2 & -- & --\\
        CoCa~\cite{yu2022coca} & \makecell[l]{\jftThreeB + \\ \alignOriginal} & \multicolumn{2}{c}{\CocaModel} & 576 & \textbf{86.3} & -- & --\\
        \midrule
        \oursShort & \igSizeShort & \vitH & ? & 224 & 80.8 & 95.8 & -- \fixme{!!} \\ %
        \oursShort & \igSizeShort & \vitTwoB & ? & 224 & 82.1 & \textbf{96.2} & -- \\ %
    \end{tabu}%
    }
    \caption{\textbf{Zero shot image classification results.} We evaluate zero-shot transfer on \inetOneKShort and \food. 
    Our models push the \sota on \foodShort, while being competitive on \inetOneKShort. 
    The best performing models on \inetOneKShort train on \jftThreeB and ALIGN, and the performance on the two datasets is not
    well correlated, exemplifying the impact of pretraining dataset choice on zero-shot transfer performance.
    }
    \label{tab:zero_shot}
\end{table}

    \end{minipage}
\end{table*}

{\bf \noindent \inetOneK image classification.}
\cref{tab:image_classification} shows the performance of different methods on \inetOneKShort.
\oursShort gets the best performance for a \vitH sized model (89.3\%).
Recent methods such as \scaleViT~\cite{zhai2022scaling} are better on \inetOneKShort and we hypothesize that this gap stems mainly from the differences in the pretraining datasets (\igSizeShort \vs \jftThreeB).
We also compute linear performance using frozen features on \inetOneKShort at 224px resolution. Our models produce strong
features which outperform other methods with \ce objectives. They also surpass the performance of the self-supervised 
\dinovTwo~\cite{oquab2023dinov2} model optimized to produce strong frozen representations:

\begin{center}
    \resizebox{0.85\linewidth}{!}{
        \begin{tabu}{llcH|c}
            \bf Method & \bf Dataset & \bf Arch. & \bf Res. & \bf \inetOneKShort Linear \\
            \midrule
            \swag~\cite{singh2022revisiting} & IG-3.6B & \vitH & 224 & 85.8 \\
            \openclip~\cite{ilharco_gabriel_2021_5143773} & \laionTwoB & \vitG & 224 & 86.2 \\
            \dinovTwo~\cite{oquab2023dinov2} & LVD-142M & ViT-g & 224 & 86.5 \\
            \midrule
            \oursShort & \igSizeShort & \vitH & 224 & 87.0 \\
            \oursShort & \igSizeShort & \vitTwoB & 224 & 88.1 \\ 
            \oursShort & \igSizeShort & \vitSixB & 224 & \bf 88.6 \\ 
        \end{tabu}
    }
\end{center}

{\bf \noindent Robustness for image classification.}
We evaluate the robustness of our models finetuned on \inetOneKShort on
additional test sets whose classes overlap with \inetOneKShort in~\cref{tab:image_classification} to evaluate the generalization
and robustness of our models to additional test sets.
We find that, despite \oursShort being 0.4\% behind \scaleViT on \inetOneKShort, it is significantly
more robust and generalizes better
on these additional test sets -- \oursShort gets the highest reported performance on \inetvTwo, \inetReal and \objectNet for \inetOneKShort
finetuned models. We also see the benefits of scaling models even up to 6.5B parameters on \inetvTwo and \objectNet, where
the performance continues to improve significantly with increases in model size.

{\bf \noindent Generalization in image classification.}
We evaluate the generalization of our model on additional fine-grained image classification using \inat.
\inatShort is a challenging long-tailed and fine-grained dataset with images of multiple species of visually similar plants and animals.
For the same size, our \vitH outperforms the previous best result~\cite{he2021masked} by 3.7\%.
Our \vitSixB sets a new \sota result on \inatShort (+4.9\% over~\cite{he2021masked})

{\bf \noindent Video classification.}
In~\cref{tab:video_classification} we investigate how \oursShort's pretraining transfers to video action classification on \kineticsShort and \sthsthShort. \oursShort is competitive with \sota
methods, including ones that pretrain on videos, whereas our models are only pretrained on images. Specifically,
our \vitL gets the highest reported performance on both video datasets.
For all video finetuning, we use relative position 
embeddings~\cite{fan2021multiscale}, which improves our performance by 0.6\% on \kineticsShort for a \vitL.
Overall, the results indicate the promise of \mae \prept for building strong video understanding models.

{\bf \noindent Low-shot image classification.}
We evaluate the label efficiency of our models using a few examples per class for finetuning.
We use two datasets, \inetOneKShort and \inatShort, with $K$ shots (labeled examples per class), $K \in \{1, 5, 10\}$.
For \inatShort, as some classes have less than $K$ images, we adapt our setting to consider \textit{at most} $K$ shots.
For each value of $K$, we generate 5 splits of the original dataset using 5 different random seeds and report 
the mean top-1 accuracy.

We evaluate two protocols for low-shot finetuning -- linear classifiers and Adapters~\cite{houlsby2019parameter}, both 
of which keep the entire model parameters frozen and introduce a few trainable parameters.
We evaluated multiple Adapters proposed for ViTs -- LoRA~\cite{hu2022lora}, 
AdaptFormer~\cite{chen2022adaptformer}, and VPT~\cite{jia2022visual}. We found that VPT performed the best while being 
robust to the choice of hyperparameters, and outperforms linear classifiers for our models.
For other works, we report with the best protocol. Full details in \cref{app:transfer_details}.

\cref{tab:low_shot_image_classification} shows a comparison with \sota methods on low-shot \inetOneKShort and 
\inatShort, including foundational and self-supervised models.
Our models show impressive low-shot performance on both \inetOneKShort and \inatShort, reaching 84.6\% and 80.9\%
top-1 accuracy with only 10 labeled examples per class, respectively. They reach the highest reported performance with just one 
labeled example per class on \inetOneKShort of 63.6\%.

{\bf \noindent Zero-shot transfer.}
Strong foundational models are expected to also have a good open world understanding of visual concepts.
To equip our pretrained vision encoders with such capabilities, we utilize \lit~\cite{zhai2022lit}. 
We initialize the text encoder from an \xlmr Large~\cite{conneau2020unsupervised} model.
\cref{tab:zero_shot} shows the zero-shot
transfer performance of our models on \inetOneK, and \food. Our \vitTwoB attains 82.1\% accuracy on 
\inetOneKShort, outperforming a similarly sized \openclip model. Our results lag behind other works which pretrain 
on datasets such as \jftThreeB and ALIGN, highlighting the importance of the pretraining dataset for performance. 
This data-advantage is also observed in the finetuning results discussed in \cref{tab:image_classification}, where 
\jftThreeB provides best \inetOneKShort accuracy. On \food we attain the highest zero-shot transfer accuracy of 96.2\%. 
The performance on the two datasets is not well correlated, further demonstrating the impact the pretraining 
dataset can have on a particular zero-shot task.

\begin{table}[t]
    \centering
    \setlength{\tabcolsep}{3pt}
    \resizebox{0.95\linewidth}{!}{
        \begin{tabu}{llcc|cc}
            \centering
            \bf Method & \bf Dataset & \bf Arch. & \bf Framework & \bf AP\textsuperscript{box} & \bf AP\textsuperscript{mask} \\
            \midrule
            \multicolumn{4}{l}{\em Results with a different detection framework} &  \\
            Winner 2021 \cite{fu2021lvis} & \inetFullShort & CBNetv2 & HTC & -- & 49.2 \\
            \midrule
            \mae \cite{li2022exploring} & \inetOneKShort & \vitDet-H & Cascade  & 51.5 & 46.6 \\
            \swag~\cite{singh2022revisiting} & IG-3.6B & \vitDet-H & Cascade  & 47.1 & 42.1  \\    %
            \midrule
            \oursShort & \igSizeShort & \vitDet-H & Cascade  & 50.8 & 45.5 \\ %
            \oursShort & \igSizeShort & \vitDet-2B & Cascade  & 51.8 & 46.1 \\ %
        \end{tabu}%
    }
    \caption{\textbf{Detection and Segmentation results on \lvisShort} (v1 val). 
    We compare \oursShort with prior work and report the detection (AP\textsuperscript{box}) and instance segmentation performance (AP\textsuperscript{mask}).
    Our \vitH significantly outperforms \swag which uses the same pretraining and model size, but without \prept.
    Our detection performance also improves with the larger 2B parameter model.
    }
    \label{tab:detection_segmentation_lvis}
\end{table}

\begin{table}[t]
    \centering
    \setlength{\tabcolsep}{3pt}
    \resizebox{\linewidth}{!}{
        \begin{tabu}{llcc|cc}
            \centering
            \bf Method & \bf Dataset & \bf Arch. & \bf Framework & \bf AP\textsuperscript{box} & \bf AP\textsuperscript{mask} \\
            \midrule
            \multicolumn{4}{l}{\em Results with different detection framework} \\
            CBNetv2 \cite{liang2022cbnet} & \inetFullShort & 2 \texttimes Swin-L & HTC & 59.1 & 51.0 \\
            SwinV2-L \cite{liu2022swin} & \inetFullShort & SwinV2-L & HTC++ & 58.9 & 51.2 \\
            Co-DETR \cite{zong2022detrs} & \inetFullShort & Swin-L & Co-DETR & 58.5 & -- \\
            \midrule
            \multicolumn{4}{l}{\em Methods that pretrain on a detection dataset} \\
            Florence \cite{yuan2021florence} & \makecell[l]{\florenceDataset \\ + FLOD-9M\textsuperscript{\textdagger}} & CoSwin-H & DyHead \cite{dai2021dynamic} & 62.0 & -- \\
            DINO \cite{zhang2022dino} & \inetFullShort + \objectDetshort\textsuperscript{\textdagger} & \swinL & -- & 63.2 & -- \\
            FocalNet \cite{yang2022focal} & \inetFullShort + \objectDetshort\textsuperscript{\textdagger} & FocalNet-H & DINO \cite{zhang2022dino} & 64.2 & -- \\
            Co-DETR \cite{zong2022detrs} & \inetFullShort + \objectDetshort\textsuperscript{\textdagger} & MixMIM-g & Co-DETR & 64.4 & -- \\
            \midrule
            \mae \cite{li2022exploring} & \inetOneKShort & \vitDet-H & Cascade & 58.7 & 50.9 \\  %
            \swag~\cite{singh2022revisiting} & IG-3.6B & \vitDet-H & Cascade  & 55.7 & 47.9 \\  %
            \midrule
            \oursShort & \igSizeShort & \vitDet-H & Cascade  & 57.7 & 49.6 \\  %
            \oursShort & \igSizeShort & \vitDet-2B & Cascade  & 58.0 & 50.1 \\  %
        \end{tabu}%
    }
    \caption{\textbf{Detection and Segmentation on \cocoShort} (val2017).
    We report the detection (AP\textsuperscript{box}) and instance segmentation performance (AP\textsuperscript{mask}).
    Our models outperform \swag which uses the same pretraining, but without \prept.
    \textsuperscript{\textdagger}Large scale detection datasets -- using additional detection data like \objectsThreeSixFive has been shown to boost performance
    by 5.6 mAP~\cite{shao2019objects365}.
    }
    \label{tab:detection_segmentation_coco}
\end{table}

{\bf \noindent Detection and segmentation.}
Next, we evaluate models on detection and instance segmentation, on the \lvisShort~\cite{gupta2019lvis} (\cref{tab:detection_segmentation_lvis}) 
and \cocoShort~\cite{lin2014microsoft} datasets (\cref{tab:detection_segmentation_coco}).
We use the Cascade Mask R-CNN framework~\cite{he2017maskrcnn}, with the \vitDet~\cite{li2022vitdet} architecture, and 
initialize the backbone with our pretrained models.
For finetuning our models, we start with the hyperparameters from~\cite{li2022vitdet} and adapt them for our models, 
full details are in~\cref{app:transfer_details}.
We also perform a system level comparison with other \sota works, but note that drawing meaningful conclusions out of this 
is difficult, on account
of the multiple differences in the detection frameworks, model architectures, and datasets.  

On both benchmarks, \oursShort considerably outperforms the weakly supervised \swag~\cite{singh2022revisiting} model, demonstrating the benefit of our 
additional \prept stage. On the long-tailed \lvisShort dataset, \oursShort outperforms \mae \inetOneKShort pretraining on detection 
AP~\cite{li2022vitdet}, 
but lags slightly behind on \cocoShort. \mae's strong performance on detection using both \inetOneKShort 
and \igSizeShort (\cref{fig:mae_scaling}) can potentially be explained by the fact that it is trained to reproduce images with 75\% 
masking, whereas \ce is only trained to predict one or a few salient objects in an image.
Lastly, methods which use additional detection data, such as \objectsThreeSixFive~\cite{shao2019objects365} or 
FLOD-9M~\cite{yuan2021florence}, have strong detection performance.

\begin{figure}[t]
\begin{center}
  \captionsetup{type=figure}
    \begin{subfigure}{.22\textwidth}
        \begin{tikzpicture}
    \begin{axis}[
        legend pos=south east,
        xmin=0,
        xmode=log,
        grid=both,
        grid style={line width=.1pt, draw=gray!10},
        major grid style={line width=.2pt,draw=gray!50},
        minor tick num=2,
        log ticks with fixed point,
        xtick={0.1, 0.5, 2.0},
        ytick={44, 49, 54},
        axis x line*=bottom,
        axis y line*=left,
        height=1.5\linewidth,
        width=1.3\linewidth,
        title style= {align=center, font=\small},
        ylabel style= {align=center, font=\small},
        xlabel style = {font=\small},
        title={\color{DetDark}{\lvisShort (AP\textsuperscript{box})}},
        xlabel={Model parameters (billions)},
        yticklabel style = {font=\small},
        xticklabel style = {font=\small},
        legend style={cells={align=left}, font=\footnotesize, fill=none, draw=none},
    ]
    \addplot[mark=*, colMAECE, very thick, mark options={solid}] plot coordinates {
        (0.086, 44.2)  %
        (0.307, 49.6)  %
        (0.632, 50.8)  %
        (1.9, 51.8)    %
    };
    \addlegendentry{\ours}
    \addplot[mark=o, colCE, very thick, mark options={solid}] plot coordinates {
        (0.086, 41.4)  %
        (0.307, 47.1)  %
        (0.632, 46.9)  %
        (1.9, 47.4)    %
    };
    \addlegendentry{\ce}
    \end{axis}
\end{tikzpicture}
    \end{subfigure} \hfill
    \begin{subfigure}{.22\textwidth}
        \begin{tikzpicture}
    \begin{axis}[
        legend pos=south east,
        xmin=0,
        xmode=log,
        grid=both,
        grid style={line width=.1pt, draw=gray!10},
        major grid style={line width=.2pt,draw=gray!50},
        minor tick num=2,
        log ticks with fixed point,
        xtick={0.1, 0.5, 2.0},
        ytick={54, 57, 60},
        axis x line*=bottom,
        axis y line*=left,
        height=1.5\linewidth,
        width=1.3\linewidth,
        title style= {align=center, font=\small},
        ylabel style= {align=center, font=\small},
        xlabel style = {font=\small},
        title={\color{DetDark}{\cocoShort (AP\textsuperscript{box})}},
        xlabel={Model parameters (billions)},
        yticklabel style = {font=\small},
        xticklabel style = {font=\small},
        legend style={cells={align=left}, font=\footnotesize, fill=none, draw=none},
    ]
    \addplot[mark=*, colMAECE, very thick, mark options={solid}] plot coordinates {
        (0.086, 53.4)  %
        (0.307, 57.3)  %
        (0.632, 57.7)  %
        (1.9, 58.0)    %
    };
    \addlegendentry{\ours}
    \addplot[mark=o, colCE, very thick, mark options={solid}] plot coordinates {
        (0.086, 51.9)  %
        (0.307, 54.9)  %
        (0.632, 55.2)  %
        (1.9, 54.8)    %
    };
    \addlegendentry{\ce}
    \end{axis}
\end{tikzpicture}
    \end{subfigure} \hfill
    \ifarxiv
        \vspace{-0.23in}
    \else
        \vspace{-0.4in}
    \fi
    \caption{\textbf{\Prept with MAE significantly boosts performance on detection}.
    Scaling the model size with \ce only does not improve performance on detection.
    Adding \mae as pre-pretraining helps \ours scaling on detection.}
    \label{fig:ce_vs_maece_detection}
\end{center}
\end{figure}
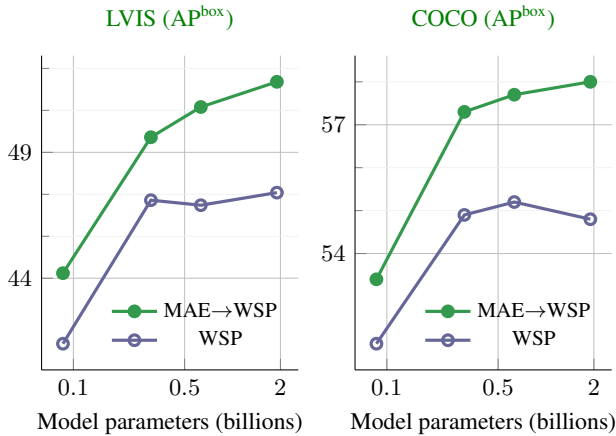

\begin{figure}[]
  \begin{center}
    \captionsetup{type=figure}
      \begin{tikzpicture}
    \begin{groupplot}[
        group style={group size= 2 by 1, vertical sep=45pt},
        legend style={cells={align=center}, font=\small, fill=none, draw=none},
    ]

        \nextgroupplot[
            ybar,
            width=0.6\linewidth,
            height=0.7\linewidth,
            ylabel style= {align=center, font=\small},
            title style= {align=center, font=\small},
            xlabel style = {font=\small},
            yticklabel style = {font=\small},
            xticklabel style = {font=\small},
            axis x line*=bottom,
            axis y line*=left,
            ymajorgrids=true,
            yminorgrids=true,
            grid style={line width=.1pt, draw=gray!10},
            major grid style={line width=.2pt,draw=gray!50},
            legend style={cells={align=left}, font=\footnotesize, fill=none, draw=none},
            enlargelimits=0.22, %
            legend style={anchor=north,legend columns=-1},
            bar width=6pt,
            symbolic x coords={
                Small,
                Medium,
                Large,
            },
            title={AP\textsuperscript{box} by object size},
            xtick=data,
        ]
            \addplot [fill=colCE, draw=colCE] coordinates {
                (Small, 34.8)
                (Medium, 57.7)
                (Large, 67.2)
            };
            \addplot [fill=colMAE, draw=colMAE] coordinates {
                (Small, 39.4)
                (Medium, 58.8)
                (Large, 68.1)
            };
            \addplot [fill=colMAECE, draw=colMAECE] coordinates {
                (Small, 37.9)
                (Medium, 60.6)
                (Large, 69.3)
            };
            \coordinate (top) at (rel axis cs:1,1); %

        \nextgroupplot[
            ybar,
            width=0.6\linewidth,
            height=0.7\linewidth,
            ylabel style= {align=center, font=\small},
            title style= {align=center, font=\small},
            xlabel style = {font=\small},
            yticklabel style = {font=\small},
            xticklabel style = {font=\small},
            axis x line*=bottom,
            axis y line*=left,
            ymajorgrids=true,
            yminorgrids=true,
            grid style={line width=.1pt, draw=gray!10},
            major grid style={line width=.2pt,draw=gray!50},
            legend style={cells={align=left}, font=\footnotesize, fill=none, draw=none},
            enlargelimits=0.22, %
            legend style={anchor=north,legend columns=-1},
            legend image code/.code={\draw [#1] (0cm,0cm) rectangle (0.5cm,0.08cm);},
            bar width=6pt,
            symbolic x coords={
                Rare,
                Common,
                Freq,
            },
            title={AP\textsuperscript{box} by class frequency},
            xtick=data,
            legend to name=zelda,
        ]
            \addplot [fill=colCE, draw=colCE] coordinates {
                (Rare, 38.2)
                (Common, 47.0)
                (Freq, 51.1)
            };
            \addlegendentry{\ce};
            \addplot [fill=colMAE, draw=colMAE] coordinates {
                (Rare, 36.4)
                (Common, 48.6)
                (Freq, 55.4)
            };
            \addlegendentry{\mae};
            \addplot [fill=colMAECE, draw=colMAECE] coordinates {
                (Rare, 38.0)
                (Common, 50.4)
                (Freq, 54.0)
            };
            \addlegendentry{\ours};

    \end{groupplot}
    \node[below] at (top |- current bounding box.south){\pgfplotslegendfromname{zelda}};
\end{tikzpicture}
      \ifarxiv
        \vspace{-0.12in}
      \else
        \vspace{-0.12in}
      \fi
      \caption{\textbf{Dissecting detection performance.}
        \lvisShort performance based on the size of objects (\textbf{left}), or the
        class occurrence frequency (\textbf{right}). \mae is stronger than \ce at detecting small objects and 
        frequent classes, and is worse on rare classes. \ours outperforms \ce on all axes. It
        is much closer to \mae on smaller objects and rare classes, and also outperforms \mae elsewhere.
      }
      \label{fig:detection_analysis_all}
  \end{center}
\end{figure}
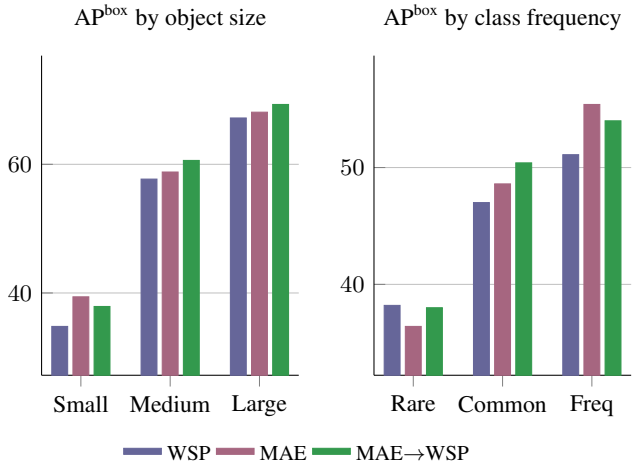

\noindent \textbf{Analyzing detection performance.}
We further inspect the benefits of \prept for detection in~\cref{fig:ce_vs_maece_detection}.
Unlike other tasks like image / video classification, scaling model size using \ce pretraining does \emph{not} improve 
detection performance.
However, adding \mae \prept provides consistent gains and allows \ce to scale with model size.

We dissect the performance of \vitL models trained with \mae, \ce, and \ours in \cref{fig:detection_analysis_all}
based on the size of the object, or by the frequency of the object's class.
We observe that \ours performs better than \ce on all tasks -- detecting rare to frequent classes across small to large object sizes.
It improves over \mae at detecting rare objects, presumably because of the diversity in the \igSizeShort labels.

\noindent \textbf{Performance over $\mathbf{\sim100\times}$ model size spectrum.} So far we have focused our evaluations on large scale models 
(\vitH, \vitTwoB, \vitSixB). In \cref{tab:overall_in1k_scaling_numbers} we show the performance of
all our models on \inetOneKShort, evaluated using linear classifiers as well as with
finetuning. We note that our approach works very well for small scale models (\vitB, \vitL) as well. 
\oursShort is 
within 0.3\% of the best performance for a \vitB (86.4\% \vs 86.7\% for \cite{Touvron2022DeiTIR}), and achieves the best performance for \vitL 
(88.8\%) on \inetOneKShort.

\begin{table}[H]
    \centering
    \setlength{\tabcolsep}{3pt}
    \resizebox{0.85\linewidth}{!}{
    \begin{tabu}{c|ccccc}
        \centering
        \bf Method & \bf \vitB & \bf \vitL & \bf \vitH & \bf \vitTwoB & \bf \vitSixB \\
        \midrule
        \multicolumn{6}{l}{\emph{\mae pretrained on \igSizeShort}} \\
        Linear 224px & 56.5 & 65.1 & 69.6 & 76.1 & 78.2 \\
        Finetune 224px & 83.5 & 86.1 & 87.4 & 87.8 & 88.3 \\
        \midrule
        \multicolumn{6}{l}{\emph{\oursShort (\ours) pretrained on \igSizeShort}} \\
        Linear 224px & 82.8 & 86.0 & 87.0 & 88.1 & 88.6 \\
        Finetune 518px & 86.4 & 88.8 & 89.3 & 89.7 & 90.1 \\
    \end{tabu}%
    }
    \caption{
        \textbf{\mae and \oursShort \inetOneKShort performance across all model scales.} We show the linear and finetuned performance on \inetOneK for \mae and \oursShort pretraining 
        on \igSize. Both \mae \prept and \oursShort scale well, starting from 86M parameters, up to 6.5B parameters.
        Using frozen features and when finetuned, \oursShort models are some of the strongest models on \inetOneK at all size scales.
    }
    \label{tab:overall_in1k_scaling_numbers}
\end{table}

\section{Conclusion}
\label{sec:conclusion}

We introduced \prept which is an initial stage in the standard pretrain-then-finetune paradigm.
\Prept uses \mae and thus, does not need additional supervision and can be conveniently added to web-scale training pipelines.
We show that \prept improves downstream performance on multiple different recognition tasks, improves model convergence, and is overall more efficient than standard weakly-supervised pretraining.
Our self-supervised \prept improves results for models trained with billions of labels, showing that it is a scalable technique that matters even at web-scale.
The benefits of using \prept hold across varying model sizes, and different pretraining data distributions showing that it is a robust technique.
Finally, \prept naturally and successfully combines the two most common pretraining strategies -- self-supervised and weakly-supervised learning.
Our results suggest that model initialization plays a significant role in the final performance, training dynamics \etc even for web-scale training with billions of parameter updates and labels, and should be further investigated.

\section*{Acknowledgements}
\label{sec:acknowledgements}

\noindent We are grateful to Kaiming He for invaluable discussions and suggestions, and his advice around \mae \prept .
We thank Mary Williamson for her help, guidance and support in project planning, execution, and managing various uncertainties throughout
the research project. 
We thank Devi Parikh for her support around the final parts of the project.
We are grateful to Vivek Pai for his help with the training infrastructure. We also thank Yanghao Li for his help with the detection 
evaluations, Dominik Kallusky for his help with the dataset pipeline, and Tsung-Yu Lin for his help with preparing the image-caption dataset. 
Lastly, we thank Stephen Roller and Naman Goyal for helpful discussions and feedback.

\clearpage

{\small
\bibliographystyle{ieee_fullname}
\bibliography{refs}
}

\clearpage
\appendix
\section*{Appendix}
\label{sec:appendix}

\section{Pretraining details}
\label{app:pretraining_details}

{\bf \noindent \mae pretraining.} We make no changes from He \etal~\cite{he2021masked} for \mae pretraining.
We utilize the same decoder dimensions as well -- 8 layers, 512 dimension, and 16 heads. Training hyperparameters are shared in
\cref{tab:mae_pretrain_settings}.

\begin{table}[!htb]
    \begin{center}
        \centering
        \resizebox{0.8\linewidth}{!}{
            \begin{tabular}{l|c}
                \bf Setting & \bf Value  \\
                \midrule
                Epochs & 1 \\
                Batch size & 4096 \\
                Masking ratio & 0.75 \\
                Optimizer & AdamW~\cite{loshchilov2017decoupled} \\
                Learning rate: \\
                \quad Schedule & Cosine \\
                \quad Peak & 2.4e-3 \\
                \quad Warmup Schedule & Linear \\
                \quad Warmup Fraction & 5\% \\
                Weight decay & 0.05 \\
                Optimizer Momentum & $\beta_1=0.9,\beta_2=0.95$ \\
                Augmentations: \\
                \quad {\tt RandomResizedCrop} \\
                \qquad {\tt size} & 224px \\
                \qquad {\tt scale} & [0.2, 1.00] \\
                \qquad {\tt ratio} & [0.75, 1.33] \\
                \qquad {\tt interpolation} & Bicubic \\
                \quad {\tt RandomHorizontalFlip} & $p=0.5$ \\
                \quad {\tt Normalize} \\
            \end{tabular}
        }
    \end{center}
    \vspace{-0.15in}
    \caption{
        \textbf{\mae Pretraining hyperparameters.} We follow the settings from \cite{he2021masked} without any modifications.
    }
    \label{tab:mae_pretrain_settings}
\end{table}

\begin{table}[!htb]
    \begin{center}
        \centering
        \resizebox{0.8\linewidth}{!}{
            \begin{tabular}{l|c}
                \bf Setting & \bf Value  \\
                \midrule
                Epochs & 1 \\
                Batch size & 8192 \\
                Optimizer & AdamW~\cite{loshchilov2017decoupled} \\
                Learning rate: \\
                \quad Schedule & Linear \\
                \quad Peak & 4e-4 \\
                \quad Warmup Schedule & Linear \\
                \quad Warmup Fraction & 5\% \\
                Weight decay & 0.1 \\
                Optimizer Momentum & $\beta_1=0.9,\beta_2=0.999$ \\
                Augmentations: \\
                \quad {\tt RandomResizedCrop} \\
                \qquad {\tt size} & 224px \\
                \qquad {\tt scale} & [0.08, 1.00] \\
                \qquad {\tt ratio} & [0.75, 1.33] \\
                \qquad {\tt interpolation} & Bicubic \\
                \quad {\tt RandomHorizontalFlip} & $p=0.5$ \\
                \quad {\tt Normalize} \\
            \end{tabular}
        }
    \end{center}
    \vspace{-0.15in}
    \caption{
        \textbf{\ce Pretraining hyperparameters.} We follow the settings from \cite{singh2022revisiting} without any modifications.
        For \vitTwoB we were able to train with the same hyperparameters, but ultimately reduced the learning rate to 1e-4, enabled
        gradient clipping to 1.0 norm, and set AdamW's $\beta_2$ to 0.95 for improved training stability.
    }
    \label{tab:ce_pretrain_settings}
\end{table}

\begin{table}[!htb]
    \begin{center}
        \centering
        \resizebox{0.8\linewidth}{!}{
            \begin{tabular}{l|c}
                \bf Setting & \bf Value  \\
                \midrule
                Epochs & 1 \\
                Batch size & 32768 \\
                Optimizer & AdamW~\cite{loshchilov2017decoupled} \\
                Learning rate: \\
                \quad Schedule & Cosine \\
                \quad Peak & 2e-4 \\
                \quad Warmup Schedule & Linear \\
                \quad Warmup Fraction & 4\% \\
                Weight decay & 0.1 \\
                Optimizer Momentum & $\beta_1=0.9,\beta_2=0.98$ \\
                Loss: \\
                \quad {\tt CLIP~\cite{radford2021learning}} \\
                \quad {\tt LabelSmoothing}~\cite{szegedy2016rethinking} & 0.1 \\
                Augmentations: \\
                \quad {\tt RandomResizedCrop} \\
                \qquad {\tt size} & 224px \\
                \qquad {\tt scale} & [0.9, 1.00] \\
                \qquad {\tt ratio} & [0.75, 1.33] \\
                \qquad {\tt interpolation} & Bicubic \\
                \quad {\tt RandomHorizontalFlip} & $p=0.5$ \\
                \quad {\tt Normalize} \\
            \end{tabular}
        }
    \vspace{-0.15in}
    \end{center}
    \caption{
        \textbf{\lit training hyperparameters} for \cref{tab:zero_shot}. 
        For ablations with \xlmr Base we used a higher learning rate of 1e-3 with a
        dropout of 0.1.
    }
    \label{tab:lit_settings}
\end{table}

\begin{table}[!htb]
    \begin{center}
        \centering
        \resizebox{0.8\linewidth}{!}{
            \begin{tabular}{l|c}
                \bf Setting & \bf Value  \\
                \midrule
                Epochs & 90 \\
                Batch size & 4096 \\
                Optimizer & AdamW~\cite{loshchilov2017decoupled} \\
                Learning rate: \\
                \quad Schedule & Cosine \\
                \quad Peak & 1e-3 \\
                \quad Warmup Schedule & Linear \\
                \quad Warmup Fraction & 3.3\% \\
                Weight decay & 0.1 \\
                Optimizer Momentum & $\beta_1=0.9,\beta_2=0.999$ \\
                Gradient Clipping & 1.0 \\
                Augmentations: \\
                \quad {\tt RandomResizedCrop} \\
                \qquad {\tt size} & 224px \\
                \qquad {\tt scale} & [0.08, 1.00] \\
                \qquad {\tt ratio} & [0.75, 1.33] \\
                \qquad {\tt interpolation} & Bicubic \\
                \quad {\tt RandomHorizontalFlip} & $p=0.5$ \\
                \quad {\tt RandomAugment~\cite{cubuk2020randaugment}} \\
                \qquad {\tt num\_layers} & 2 \\
                \qquad {\tt magnitude} & 9 \\
                \quad {\tt Normalize} \\
                \quad {\tt mixup~\cite{zhang2017mixup}} & 0.5 \\
            \end{tabular}
        }
    \end{center}
    \vspace{-0.15in}
    \caption{
        \textbf{\ce Pretraining hyperparameters for \inetFullShort.}
    }
    \label{tab:pretrain_details_inet21k}
\end{table}

{\bf \noindent \ce and \ours pretraining.}
We note that large scale \ce pretraining is quite robust to the choice of training hyperparameters, including the choice of
(i) a single \vs two layer MLP head, (ii) softmax cross-entropy \vs binary cross-entropy training loss, (iii) additional training
augmentations like mixup~\cite{zhang2017mixup}, CutMix~\cite{yun2019cutmix}, \etc, (iv) learning rate over a $\sim5\times$ range.
Most of these choices seem to affect results at small scales, like 0.1 epochs over \igSizeShort (500 million samples seen), but the
differences dissipate when training for a full epoch (5 billion samples seen). Our training hyperparameters are shared in
\cref{tab:ce_pretrain_settings}, and we use these settings for both \ce and \ours.
We follow Singh \etal~\cite{singh2022revisiting} for the training setup and hyperparameters for consistency and easier comparisons.
For a label vocabulary $C$, and an image $img$ with labels $L_{img} \in \{0, 1\}^{|C|}$, we utilize softmax cross-entropy,
where our label is normalized to a probability
distribution, $L_{im} / \sum_{c \in C}{L_{im}^c}$, and the model output is passed to a softmax followed by a cross-entropy loss
\cite{singh2022revisiting, mahajan2018exploring}.
We attach a two layer MLP head to the trunk -- \texttt{\small 
    Linear(embed, embed), Tanh(), Linear(embed, classes)}.

{\bf \noindent \lit training.} We follow \lit to train a text encoder on \ig captions.
We sanitize the captions and also remove the pound signs in front of hashtags.
We use a context length (number of tokens) of 100 per caption. Following \clip, we chose the embedding dimension
for aligning the encoders to be 512, 768 and 1024 for \vitB, \vitL and \vitH, respectively, and set it to 2048 for \vitTwoB. 
We use pretrained \xlmr~\cite{conneau2020unsupervised} text encoders, with the Base size (270M parameters) for ablations and
Large (550M parameters) for the results in \cref{tab:zero_shot}. Our \lit training hyperparameters are shared in 
\cref{tab:lit_settings}.

\begin{table}[!htb]
    \begin{center}
        \centering
        \resizebox{0.8\linewidth}{!}{
            \begin{tabular}{l|c}
                \bf Setting & \bf Value  \\
                \midrule
                Batch size & 1024 \\
                Optimizer & AdamW~\cite{loshchilov2017decoupled} \\
                Learning rate: \\
                \quad Schedule & Constant \\
                \quad Peak & 2e-3 \\
                \quad Layerwise decay~\cite{clark2020electra, bao2021beit} & 0.75 \\
                \quad Warmup Schedule & Linear \\
                \quad Warmup Fraction & 5\% \\
                Weight decay & 0.05 \\
                Optimizer Momentum & $\beta_1=0.9,\beta_2=0.999$ \\
                DropPath~\cite{huang2016deep} & 0.2 \\
                EMA~\cite{polyak1992acceleration} & 1e-4 \\
                Augmentations: \\
                \quad {\tt RandomResizedCrop} \\
                \qquad {\tt size} & 518px \\
                \qquad {\tt scale} & [0.08, 1.00] \\
                \qquad {\tt ratio} & [0.75, 1.33] \\
                \qquad {\tt interpolation} & Bicubic \\
                \quad {\tt RandomHorizontalFlip} & $p=0.5$ \\
                \quad {\tt Normalize} \\
                \quad {\tt mixup~\cite{zhang2017mixup}} & 0.8 \\
                \quad {\tt CutMix~\cite{yun2019cutmix}} & 1.0 \\
                \quad {\tt LabelSmoothing~\cite{szegedy2016rethinking}} & 0.1 \\
            \end{tabular}
        }
    \end{center}
    \vspace{-0.15in}
    \caption{
        \textbf{\mae Image finetuning hyperparameters.} We train for 50 epochs on \inetOneKShort and 100 epochs on
        \inatShort.
    }
    \label{tab:mae_image_ft_settings}
\end{table}

\begin{table}[!htb]
    \begin{center}
        \centering
        \resizebox{0.67\linewidth}{!}{
            \begin{tabular}{l|c}
                \bf Setting & \bf Value  \\
                \midrule
                Batch size & 2048 \\
                Optimizer & SGD \\
                Learning rate: \\
                \quad Schedule & Constant \\
                \quad Peak & 2.4e-2 \\
                \quad Warmup Schedule & Linear \\
                \quad Warmup Fraction & 5\% \\
                Weight decay & 0.0 \\
                Optimizer Momentum & 0.9 \\
                Gradient Clipping & 1.0 \\
                DropPath~\cite{huang2016deep} & 0.2 \\
                EMA~\cite{polyak1992acceleration} & 1e-4 \\
                Augmentations: \\
                \quad {\tt RandomResizedCrop} \\
                \qquad {\tt size} & 518px \\
                \qquad {\tt scale} & [0.08, 1.00] \\
                \qquad {\tt ratio} & [0.75, 1.33] \\
                \qquad {\tt interpolation} & Bicubic \\
                \quad {\tt RandomHorizontalFlip} & $p=0.5$ \\
                \quad {\tt Normalize} \\
                \quad {\tt mixup~\cite{zhang2017mixup}} & 0.1 \\
            \end{tabular}
        }
    \end{center}
    \vspace{-0.15in}
    \caption{
        \textbf{\ce Image finetuning hyperparameters.} We train for 50 epochs on \inetOneKShort and 150 epochs on
        \inatShort.
    }
    \label{tab:ce_image_ft_settings}
\end{table}

{\bf \noindent \inetFull pretraining.}
In \inetFullShort each image is labeled with one class, but the classes are based on WordNet synsets~\cite{miller1995wordnet} which are hierarchical in nature.
Some images in this dataset are duplicated across
more than one class -- we deduplicate the images by hashing the image contents, and
convert the dataset to a multi-label dataset.
We disregard the class hierarchy amongst the labels and treat them independently.

For \mae pretraining we again follow \cite{he2021masked} and use the hyperparameters in \cref{tab:mae_pretrain_settings} and train
for 160 epochs over the dataset.
For \ce (and \ours) pretraining on \inetFull, we use the hyperparameters from Steiner \etal\cite{steiner2021train}, with a few minor
differences. We train for 90 epochs over the dataset,
and select the augmentation setting \textit{medium2} of the paper. %
We train with a softmax cross-entropy loss similar to \igSizeShort pretraining, but utilize a head with just a single linear layer.
Full hyperparameters are in \cref{tab:pretrain_details_inet21k}.

For \lit finetuning of models pretrained on \inetFullShort, we follow a similar setup and hyperparameters used for \igSizeShort (\cref{tab:lit_settings}),
and train for 20 epochs on \pmd~\cite{singh2022flava}.
Unlike \igSize dataset which has 5 billion image-text pairs, \pmd has only about 70 million image-text pairs, so we train for multiple epochs on this dataset.

\section{Transfer learning details}
\label{app:transfer_details}

\begin{table}[!htb]
    \begin{center}
        \centering
        \resizebox{0.75\linewidth}{!}{
            \begin{tabular}{l|cc}
                \multirow{2}{*}{\bf Setting} & \multicolumn{2}{c}{\bf Value}  \\
                \cmidrule{2-3}
                & \bf \kineticsShort & \bf \sthsthShort  \\
                \midrule
                Epochs & 110 & 100 \\
                Batch size & 256 & 512 \\
                Optimizer & \multicolumn{2}{c}{AdamW~\cite{loshchilov2017decoupled}} \\
                Learning rate: \\
                \quad Schedule & \multicolumn{2}{c}{Constant} \\
                \quad Peak & 1e-4 & 2e-3 \\
                \quad Layerwise decay~\cite{clark2020electra, bao2021beit} & \multicolumn{2}{c}{0.75} \\
                \quad Warmup Schedule & \multicolumn{2}{c}{Linear} \\
                \quad Warmup Fraction & \multicolumn{2}{c}{20\%} \\
                Weight decay & \multicolumn{2}{c}{0.05} \\
                Optimizer Momentum & \multicolumn{2}{c}{$\beta_1=0.9,\beta_2=0.999$} \\
                Gradient Clipping & \multicolumn{2}{c}{1.0} \\
                Dropout~\cite{srivastava2014dropout} & -- & 0.5 \\
                DropPath~\cite{huang2016deep} & 0.2 & 0.4 \\
                EMA~\cite{polyak1992acceleration} & \multicolumn{2}{c}{1e-4} \\
                Augmentations: \\
                \quad {\tt ShortSideScale} & \multicolumn{2}{c}{256px} \\
                \quad {\tt RandomResizedCrop} \\
                \qquad {\tt size} & \multicolumn{2}{c}{224px} \\
                \qquad {\tt scale} & \multicolumn{2}{c}{[0.08, 1.0]} \\
                \qquad {\tt ratio} & \multicolumn{2}{c}{[0.75, 1.33]} \\
                \qquad {\tt interpolation} & \multicolumn{2}{c}{Bicubic} \\
                \quad {\tt RandomAugment~\cite{cubuk2020randaugment}} \\
                \qquad {\tt magnitude} & \multicolumn{2}{c}{7} \\
                \qquad {\tt num\_layers} & \multicolumn{2}{c}{5} \\
                \quad {\tt RandomErasing~\cite{zhong2020random}} & \multicolumn{2}{c}{$p=0.25$} \\
                \quad {\tt Normalize} & \multicolumn{2}{c}{Yes} \\
                \quad {\tt mixup~\cite{zhang2017mixup}} & \multicolumn{2}{c}{0.8} \\
                \quad {\tt CutMix~\cite{yun2019cutmix}} & \multicolumn{2}{c}{1.0} \\
                \quad {\tt LabelSmoothing~\cite{szegedy2016rethinking}} & \multicolumn{2}{c}{0.1} \\
            \end{tabular}
        }
    \end{center}
    \vspace{-0.15in}
    \caption{
        \textbf{Video finetuning hyperparameters.} We used these settings for \cref{tab:video_classification}. For
        \cref{fig:mae_vs_ce_vs_mae_ce_ig} and \cref{fig:mae_vs_ce_vs_mae_ce_in21k} we used shorter 50 epoch schedules
        for efficiency reasons, which has minimal impact on performance (< 0.5\%).
    }
    \label{tab:video_ft_settings}
\end{table}

\begin{table}[!htb]
    \begin{center}
        \centering
        \resizebox{0.8\linewidth}{!}{
            \begin{tabular}{l|cc}
                \multirow{2}{*}{\bf Setting} & \multicolumn{2}{c}{\bf Value}  \\
                \cmidrule{2-3}
                & \bf 1-shot & \bf \{5, 10\}-shot  \\
                \midrule
                Epochs & 56 & 28 \\
                Batch size & \multicolumn{2}{c}{128} \\
                Optimizer & \multicolumn{2}{c}{SGD} \\
                Learning rate: \\
                \quad Schedule & \multicolumn{2}{c}{Cosine} \\
                \quad Peak & \multicolumn{2}{c}{6e-3} \\
                Weight decay & \multicolumn{2}{c}{0.0} \\
                Optimizer Momentum & \multicolumn{2}{c}{0.9} \\
                Augmentations: \\
                \quad {\tt ShortSideScale} & \multicolumn{2}{c}{256px} \\
                \quad {\tt RandomResizedCrop} \\
                \qquad {\tt size} & \multicolumn{2}{c}{224px} \\
                \qquad {\tt scale} & \multicolumn{2}{c}{[0.08, 1.0]} \\
                \qquad {\tt ratio} & \multicolumn{2}{c}{[0.75, 1.33]} \\
                \qquad {\tt interpolation} & \multicolumn{2}{c}{Bicubic} \\
                \quad {\tt RandomHorizontalFlip} & \multicolumn{2}{c}{$p=0.5$} \\
                \quad {\tt Normalize} & \multicolumn{2}{c}{Yes} \\
                \quad {\tt mixup~\cite{zhang2017mixup}} & \multicolumn{2}{c}{0.1} \\
                VPT: \\
                \quad {\tt NumTokens} & \multicolumn{2}{c}{8} \\
                \quad {\tt DimTokens} & \multicolumn{2}{c}{192} \\
                \quad {\tt TokensDropout} & \multicolumn{2}{c}{0.0} \\
            \end{tabular}
        }
    \end{center}
    \vspace{-0.15in}
    \caption{
        \textbf{Low-shot hyperparameters.} Settings used for our adapting our models with VPT~\cite{jia2022visual}.
        For \vitTwoB and \vitSixB, we decrease the learning rate to 1e-3, double the training epochs, and reduce the batch size to 64.
    }
    \label{tab:hyperparameters_lowshot}
\end{table}

\begin{table}[t]
    \centering
    \setlength{\tabcolsep}{3pt}
    \resizebox{0.75\linewidth}{!}{
        \begin{tabu}{l|Hccc|c}
            \centering
            \multirow{2}{*}{\bf Setting} & \multicolumn{4}{c|}{\bf \ce} & \bf \mae \\
            \cmidrule{2-6}
            & \bf \vitB & \bf \vitL & \bf \vitH & \bf \vitTwoB & \bf \vitTwoB \\
            \midrule
            Epochs & 100 & 100 & 100 & 100 & 100 \\
            Image Size (px) & 1024 & 1024 & 1024 & 1024 & 1024 \\
            Learning rate \\
            \quad Peak & 1e-4 & 1e-4 & 1e-4 & 1e-4 & 2e-4 \\
            \quad Schedule & Step & Step & Step & Step & Step \\
            \quad Layer Decay & 0.7 & 0.8 & 0.85 & 0.8 & 0.8 \\
            \quad Min Layer Decay & -- & 0.1 & 0.1 & 0.1 & -- \\
            Optimizer \\
            \quad AdamW $\beta_1$ & 0.9 & 0.9 & 0.9  & 0.9 & 0.9 \\
            \quad AdamW $\beta_2$ & 0.999 & 0.999 & 0.998 & 0.999 & 0.999 \\
            Drop Path rate & 0.1 & 0.4 & 0.4 & 0.4 & 0.5 \\
            Weight Decay & 0.1 & 0.2 & 0.1 & 0.1 & 0.1 \\
        \end{tabu}%
    }
    \caption{\textbf{Detection parameters used for \lvisShort}.
    For \mae on \vitDet architectures smaller than \vitH, we use the best set of hyper-parameters reported in the original \vitDet~\cite{li2022vitdet} work.
    }
    \label{tab:detection_parameters_lvis}
\end{table}

\begin{table}[t]
    \centering
    \setlength{\tabcolsep}{3pt}
    \resizebox{0.75\linewidth}{!}{
        \begin{tabu}{l|Hccc|c}
            \centering
            \multirow{2}{*}{\bf Setting} & \multicolumn{4}{c|}{\bf \ce} & \bf \mae \\
            \cmidrule{2-6}
            & \bf \vitB & \bf \vitL & \bf \vitH & \bf \vitTwoB & \bf \vitTwoB \\
            \midrule
            Epochs & 100 & 100 & 75 & 75 & 75 \\
            Image Size (px) & 1024 & 1024 & 1024 & 1024 & 1024 \\
            Learning rate \\
            \quad Peak & 1e-4 & 1e-4 & 1e-4 & 1e-4 & 1e-4 \\
            \quad Layer Decay & 0.7 & 0.8 & 0.85 & 0.8 & 0.8 \\
            \quad Min Layer Decay & - & - & 0.1 & 0.1 & - \\
            Optimizer \\
            \quad AdamW $\beta_1$ & 0.9 & 0.9 & 0.9  & 0.9 & 0.9 \\
            \quad AdamW $\beta_2$ & 0.999 & 0.999 & 0.999 & 0.999 & 0.999 \\
            Drop Path rate & 0.1 & 0.4 & 0.4 & 0.5 & 0.5 \\
            Weight Decay & 0.1 & 0.1 & 0.1 & 0.1 & 0.1 \\
        \end{tabu}%
    }
    \caption{\textbf{Detection parameters used for \cocoShort}.
    For \mae on \vitDet architectures smaller than \vitH, we use the best set of hyper-parameters reported in the original \vitDet~\cite{li2022vitdet} work.
    }
    \label{tab:detection_parameters_coco}
\end{table}

{\bf \noindent Image classification details.} We finetune models pretrained with \mae, \ce and \ours by attaching a single linear
layer on \inetOneKShort and \inatShort. We finetune models at either $224\times{}224$ resolution or $518\times{}518$ resolution (or
$512\times{}512$ for models which use a patch size of 16) and use the same hyperparameters at all resolutions.
The hyperparameters for high resolution finetuning are shared in \cref{tab:mae_image_ft_settings} for \mae models
and in \cref{tab:ce_image_ft_settings} for models pretrained with \ce or \ours.

{\bf \noindent Video classification details.}
For video finetuning, we sample 32 frames out of 2.7 second clips for \kinetics and 4 second clips for \sthsth, and train the
models at $224\times{}224$ resolution. We convert the
input into patches of size $2\times{}16\times{}16$, akin to MAE approaches applied to video
models~\cite{feichtenhofer2022masked,tong2022videomae,girdhar2023omnimae}. We initialize the video models with
weights from our inflated~\cite{carreira2017quo} pretrained image models. \cref{tab:video_ft_settings}
contains the finetuning hyperparameters for both \kineticsShort and \sthsthShort.

{\bf \noindent Low-shot image classification details.}
We adapt all \ours and \swag models with VPT~\cite{jia2022visual} with the same hyperparameters -- 
8 tokens of size 192 per self attention layer, as this proved to be a reasonable default setting.
The full settings for training our models with VPT are described in \cref{tab:hyperparameters_lowshot}.

We also attempted to train \clip and \openclip models with Adapters, but however noticed that they 
do not work as great with VPT as they do with logistic regression, despite sweeping a range of different hyperparameters.
We ultimately adopt the logistic regression protocol of \msn~\cite{assran2022masked} for \clip and \openclip, and
also for \dino and \msn.
For \mae low-shot evaluations, we noticed that neither Adapters nor logistic regression led to great results,
something already seen in \msn~\cite{assran2022masked}. 
We therefore follow \msn and finetune \mae in low-shot settings, but improve upon the results published in 
\cite{assran2022masked} for \mae significantly.
All these results are reported in \cref{tab:low_shot_image_classification}.

{\bf \noindent Zero-shot transfer details.} For evaluation of our \lit models,
we follow the zero-shot evaluation strategy proposed in \clip~\cite{radford2021learning}. We leverage the prompt templates 
and class names introduced in \clip for \inetOneK and
\food. We compute the cosine similarity between the query image and all the generated prompts for each class.
While doing so, we also take advantage of the class prompt ensembling approach introduced
in \clip by considering multiple prompt templates for each class.

{\bf \noindent Detection and segmentation details.}
We train all of our models with the Cascade Mask-RCNN~\cite{he2017maskrcnn} framework and
use the \vitDet~\cite{li2022vitdet} architecture to leverage our \vit models within this framework.

For our \mae models trained on \igShort data, we use the hyperparameters of \mae trained on \inetOneKShort from \cite{li2022vitdet}.
For the large \vitTwoB model, we use settings similar to \vitH and only change the layer decay to be the same as \vitL,
since both those models have $24$ transformer layers.

For \ce and \ours pretraining, we adapt the parameters used for \mae pretraining.
The most salient change is a modification of the layer decay scheme of \mae to cap it at a minimum value of $0.1$,
allowing \ce pretrained models to update the initial layers to align better for detection and segmentation tasks.
This was particularly useful for instance segmentation.

The hyperparameters used for training our detection and instance segmentation models are available
in \cref{tab:detection_parameters_lvis} for \lvisShort and \cref{tab:detection_parameters_coco} for \cocoShort.

\section{Additional Results and Full Tables}

\begin{table}[h]
    \centering
    \setlength{\tabcolsep}{3pt}
    \resizebox{0.8\linewidth}{!}{
    \begin{tabu}{c|cc|cccc}
        \centering
        \bf Arch. & \bf \makecell[c]{\inetOneKShort \\ Top-1} & \bf \makecell[c]{\inatShort \\ Top-1} & \bf \makecell[c]{\lvisShort \\ \bf AP\textsuperscript{box}} & \bf \makecell[c]{\lvisShort \\ AP\textsuperscript{mask}} & \bf \makecell[c]{\cocoShort \\ \bf AP\textsuperscript{box}} & \bf \makecell[c]{\cocoShort \\ AP\textsuperscript{mask}} \\
        \midrule
        \multicolumn{4}{l}{\emph{\inetOneKShort pretraining}} \\
        \vitB & 83.5& 75.0 & 43.0& 38.9 & 54.0 & 46.7 \\
        \vitL & 86.0& 80.2 & 49.2& 44.5 & 57.6 & 50.0 \\
        \vitH & 86.9& 82.8 & \bf 51.5& \bf 46.6 & \bf 58.7 & \bf 51.0 \\
        \vitTwoB & \bf 87.4 & \bf 84.5 & --\textsuperscript{\textdagger} & --\textsuperscript{\textdagger} & --\textsuperscript{\textdagger} & --\textsuperscript{\textdagger} \\
        \midrule
        \multicolumn{4}{l}{\emph{\igSizeShort pretraining}} \\
        \vitB & 83.5 & 74.7 & 42.9 & 38.8 & 53.8 & 46.5 \\
        \vitL & 86.1 & 80.7 & 49.0 & 44.5 & 58.0 & 50.3 \\
        \vitH & 87.4 & 84.0 & 52.7 & 47.5 & 59.1 & 51.2 \\
        \vitTwoB & 87.8 & 85.6 & \bf 53.6 & \bf 48.6 & \bf 59.9 & \bf 52.0\\
        \vitSixB & \bf 88.3 & \bf 86.6 & --\textsuperscript{\textasteriskcentered} & --\textsuperscript{\textasteriskcentered} & --\textsuperscript{\textasteriskcentered} & --\textsuperscript{\textasteriskcentered} \\
    \end{tabu}%
    }
    \caption{
        \textbf{Scaling \mae with model and dataset size.} We show the performance in tabular form of \mae pretraining on \inetOneK 
        and \igSize. \mae scales with both data and model size.
        \inetOneKShort and \inatShort finetuning results are at 224px.
        \textsuperscript{\textdagger}Training was unstable. \textsuperscript{\textasteriskcentered}Training skipped due to compute limitations.
    }
    \label{tab:mae_scaling_numbers}
\end{table}

\begin{table}[h]
    \centering
    \setlength{\tabcolsep}{3pt}
    \resizebox{0.9\linewidth}{!}{
    \begin{tabu}{ll|c|cccccc}
        \centering
        \bf Method & \bf Arch. & \bf AP & \bf Cls & \bf Loc & \bf Both & \bf Dupl & \bf Bkgd & \bf Miss \\
        \midrule
        \multicolumn{4}{l}{\em Results with IoU threshold of $0.5$} \\
        \ce & \vitL & 56.4 & 11.46 & 6.09 & 1.5 & 0.16 & 11.64 & 12.34 \\
        \mae & \vitL & 57.4 & 13.87 & 5.29 & 1.21 & 0.13 & 11.37 & 10.75 \\
        \ours & \vitL & 58.4 & 11.73 & 5.64 & 1.69 & 0.17 & 10.82 & 11.54 \\
        \midrule
        \multicolumn{4}{l}{\em Results with IoU threshold of $0.75$} \\
        \ce & \vitL & 45.0 & 8.24 & 17.96 & 2.02 & 0.0 & 10.76 & 16.04 \\
        \mae & \vitL & 47.5 & 10.98 & 15.57 & 1.49 & 0.0 & 10.92 & 13.52 \\
        \ours & \vitL & 47.1 & 9.15 & 17.00 & 2.12 & 0.0 & 10.47 & 13.85 \\
    \end{tabu}%
    }
    \caption{\textbf{TIDE~\cite{tide-eccv2020} analysis on LVIS} for our \mae, \ce and \ours pretrained \vitL models, 
    evaluated with two different IoU thresholds.
    The error contributions are computed with the \textit{progressive} mode of TIDE.
    Overall, \mae is stronger at finding and localisation of objects while \ce is stronger at classifying boxes.
    \ours strikes a balance between both.
    }
    \label{tab:detection_tide_all}
\end{table}

\begin{table*}[h]
    \centering
    \setlength{\tabcolsep}{3pt}
    \resizebox{0.6\linewidth}{!}{
    \begin{tabu}{l|cc|cc|c|c|cc|cc}
        \centering
        \multirow{3}{*}{\bf Method} & \multicolumn{6}{c|}{\bf Image Classification} & \multicolumn{2}{c|}{\bf Video Cls.} & \multicolumn{2}{c}{\bf Detection} \\
        \cmidrule{2-11}
        & \bf \inetOneKShort & \bf \inatShort & \makecell[c]{\bf \inetOneKShort \\ \bf 5-shot} & \makecell[c]{\bf \inatShort \\ \bf 5-shot} & \bf \makecell[c]{\inetOneKShort \\ \bf Zero-shot} & \makecell[c]{\bf \inetOneKShort \\ \bf Linear} & \bf \kineticsShort & \bf \sthsthShort & \bf \lvisShort & \bf \cocoShort \\
        \midrule
        \multicolumn{6}{l}{\emph{\igSizeShort Pretraining}} \\
        \mae
        & 87.2 %
        & 86.5 %
        & 37.2 %
        & 43.8 %
        & 46.3 %
        & 65.1 %
        & 82.7 %
        & \underline{73.8} %
        & \bf 58.0 %
        & \underline{49.0} %
        \\
        \ce
        & \underline{87.6} %
        & \underline{86.6} %
        & \underline{75.7} %
        & \underline{62.7} %
        & \underline{77.2} %
        & \underline{84.9} %
        & \underline{83.7} %
        & 69.8 %
        & 54.9 %
        & 47.1 %
        \\
        \ours
        & \bf 88.8 %
        & \bf 89.0 %
        & \bf 77.9 %
        & \bf 65.2 %
        & \bf 78.3 %
        & \bf 86.0 %
        & \bf 85.6 %
        & \bf 74.1 %
        & \underline{57.3} %
        & \bf 49.6 %
        \\
        \midrule
        \multicolumn{6}{l}{\emph{\inetFullShort Pretraining}} \\
        \mae
        & \underline{86.9} %
        & \underline{86.1} %
        & 37.1 %
        & 43.4 %
        & 41.7 %
        & 68.8 %
        & \underline{80.5} %
        & \bf 72.6 %
        & \bf 57.2 %
        & \bf 47.7 %
        \\
        \ce 
        & 84.9 %
        & 85.4 %
        & \underline{76.3} %
        & \underline{58.9} %
        & \underline{69.8} %
        & \underline{81.3} %
        & 80.2 %
        & 65.2 %
        & 53.3 %
        & 43.9 %
        \\
        \ours 
        & \bf 87.1 %
        & \bf 87.6 %
        & \bf 78.7 %
        & \bf 63.9 %
        & \bf 72.7 %
        & \bf 84.7 %
        & \bf 83.9 %
        & \underline{71.5} %
        & \underline{56.3} %
        & \underline{47.5} %
        \\
    \end{tabu}%
    }
    \caption{
        \textbf{MAE pre-pretraining improves performance.} We show in tabular form the transfer performance of a \vitL
        pretrained with \mae, \ce, and \ours using \igSizeShort (earlier shown in \cref{fig:mae_vs_ce_vs_mae_ce_ig}) and 
        \inetFullShort (earlier shown in \cref{fig:mae_vs_ce_vs_mae_ce_in21k}). We \textbf{bolden} the best results and 
        \underline{underline} the second best results for easier comparison. 
        Regardless of whether \ce or \mae performs better, \ours can match or even outperform either approaches. 
        \inetOneKShort and \inatShort finetuning results are at 512px resolution.
    }
    \label{tab:radar_numbers}
\end{table*}

\end{document}